\documentclass[11pt,a4paper]{article}
\usepackage[hyperref]{acl2021}
\usepackage{times}
\usepackage{latexsym}
\usepackage{todonotes}
\usepackage{subfigure}
\usepackage{adjustbox}
\usepackage{framed}
\usepackage{booktabs}
\usepackage{multirow}

\usepackage{microtype}

\aclfinalcopy % Uncomment this line for the final submission
\title{Membership Inference Attack Susceptibility of Clinical Language Models}

\author{Abhyuday Jagannatha$^{1}$, Bhanu Pratap Singh Rawat$^{1}$, Hong Yu$^{1,2}$ \\
$^1$College of Information and Computer Sciences,University of Massachusetts Amherst 
\\$^2$Department of Computer Science,University of Massachusetts Lowell \\
\texttt{\{abhyuday, brawat hongyu\}@cs.umass.edu}
}
\date{}

\begin{document}
\maketitle

\begin{abstract}

Deep Neural Network (DNN) models have been shown to have high empirical privacy leakages. 
Clinical language models (CLMs) trained on clinical data have been used to improve performance in biomedical natural language processing tasks. In this work, we investigate the risks of training-data leakage through 
white-box or black-box access to CLMs. We design and employ membership inference attacks to estimate the empirical privacy leaks for model architectures like \textsc{bert} and \textsc{gpt2}. We show that membership inference attacks on CLMs lead to non-trivial privacy leakages of up to 7\%. Our results show that smaller models have lower empirical privacy leakages than larger ones, and masked LMs have lower leakages than auto-regressive LMs. We further show that differentially private CLMs can have improved model utility on clinical domain while ensuring low empirical privacy leakage. Lastly, we also study the effects of group-level membership inference and disease rarity on CLM privacy leakages.

\end{abstract}

\section{Introduction}
\label{sec:intro}
Large contextual word embedding models have been proposed and used for obtaining the state-of-the-art performance on various NLP tasks \citep{wang2018glue,wang2019superglue}. The high performance and relatively lower sample complexity requirements achieved by domain-specific pretrained language models (LMs) \cite{peng2019transfer,devlin2018bert} have propelled their use in private data domains like healthcare and finance. A primary concern in deploying such models in a public setting is that models trained on sensitive private data may leak information about sensitive training samples. In application domains such as healthcare, such a leak may violate patient rights and cause potential harm to the participants of the study. 

De-identification or obfuscation of the underlying text may not necessarily mitigate all privacy concerns since anonymized data can be identified using additional sources \citep{narayanan2008robust}. Also, de-identification of large text datasets is only feasible through automated de-identification, which may lead to imperfect results. Differential Privacy (DP) \citep{dwork2014algorithmic} is a systematic approach to limiting worst-case training data leakage in the context of machine learning.  
Intuitively, DP limits training data leakage by limiting the effect of adding any random data sample to the training set. This is a general definition for estimating privacy budgets and is usually defined independently of the nature of the underlying data.  As a result, DP may be overly conservative for systems using high dimensional input with lower rank data space. For instance, in NLP where a topically relevant sentence is a very small subset of all randomly constructed sentences, mitigating the risk for any random data sample may not provide an efficient privacy mechanism. Moreover, DP based privacy accountants \citep{jayaraman2019evaluating} vastly overestimate the privacy leakages of large neural network models and may be practically unusable. 

On the other hand, empirical privacy leakage estimation may provide more accurate privacy estimates. However, training data extraction methods such as \citet{carlini2020extracting} may be model-specific and do not provide a standardized way to estimate privacy leakages across models architectures. Also, due to the high dimensional nature of inputs to LMs, inference attacks that demonstrate the potential of data leakages from such models are difficult to construct and study. This poses a problem for the study of privacy leakage in large LMs. Regulations like HIPPAA and EU 2016/679 have been enacted to ensure patient privacy rights. These laws dictate the use of patient data for secondary uses like research. However, they do not provide explicit guidelines for data leakage through release of Machine Learning or NLP model files. Due to a lack of clarity, sensitive data models are mostly used \emph{in-house}. Systematic privacy leakage studies for large NLP systems can inform and help policy decisions in this domain. This can accelerate the development of private, high utility NLP models. 

In this work, we design general membership inference attacks \citep{shokri2017membership} that can be used to quantitatively estimate the discrepancy in a large LM's response to training and out-of-training (OOT) samples. Recent work shows that this discrepancy, which contributes to the generalization gap, is related to the model's privacy leakage \citep{yeom2018privacy}. We use white-box and black-box membership inference adversaries to establish a standardized framework for our study.

Our main focus in this study is to examine the susceptibility of CLMs to membership attacks and empirically estimate their privacy leakage. We also investigate DP methods to make these modes more private. To the best of our knowledge, this is the first attempt to study empirical privacy leakages in CLMs and use differentially private CLM training. Our main findings are as follows: 
\begin{enumerate}
    \itemsep0em
    \item Large LMs can have higher empirical privacy leakages (7\%) than smaller LMs (2\%).
    \item Randomly masked LMs have lower privacy leakages than autoregressive LMs.
    \item $(\epsilon,\delta)-$DP training using DP-SGD \citep{dwork2014algorithmic} can reduce empirical privacy leakages while ensuring increased model utility. 
    \item Group-level membership inference attacks lead to higher privacy leakage than sample-level. 
    \item Patients with rarer disease profiles may be more vulnerable to higher privacy leakages. 
\end{enumerate}

\begin{figure*}
    \centering
    \subfigure[]{\includegraphics[width=0.3\textwidth]{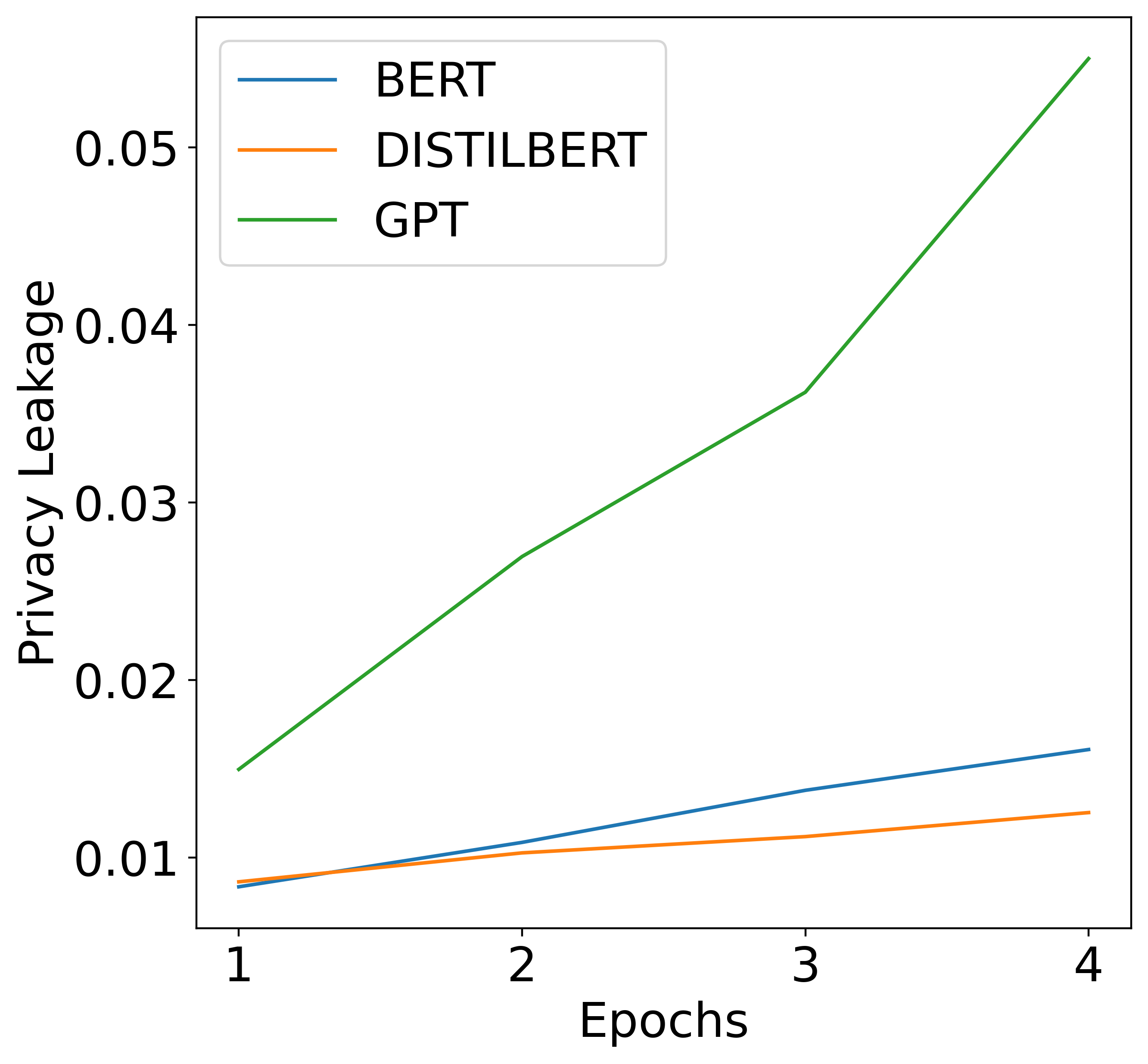}}
    \subfigure[]{\includegraphics[width=0.3\textwidth]{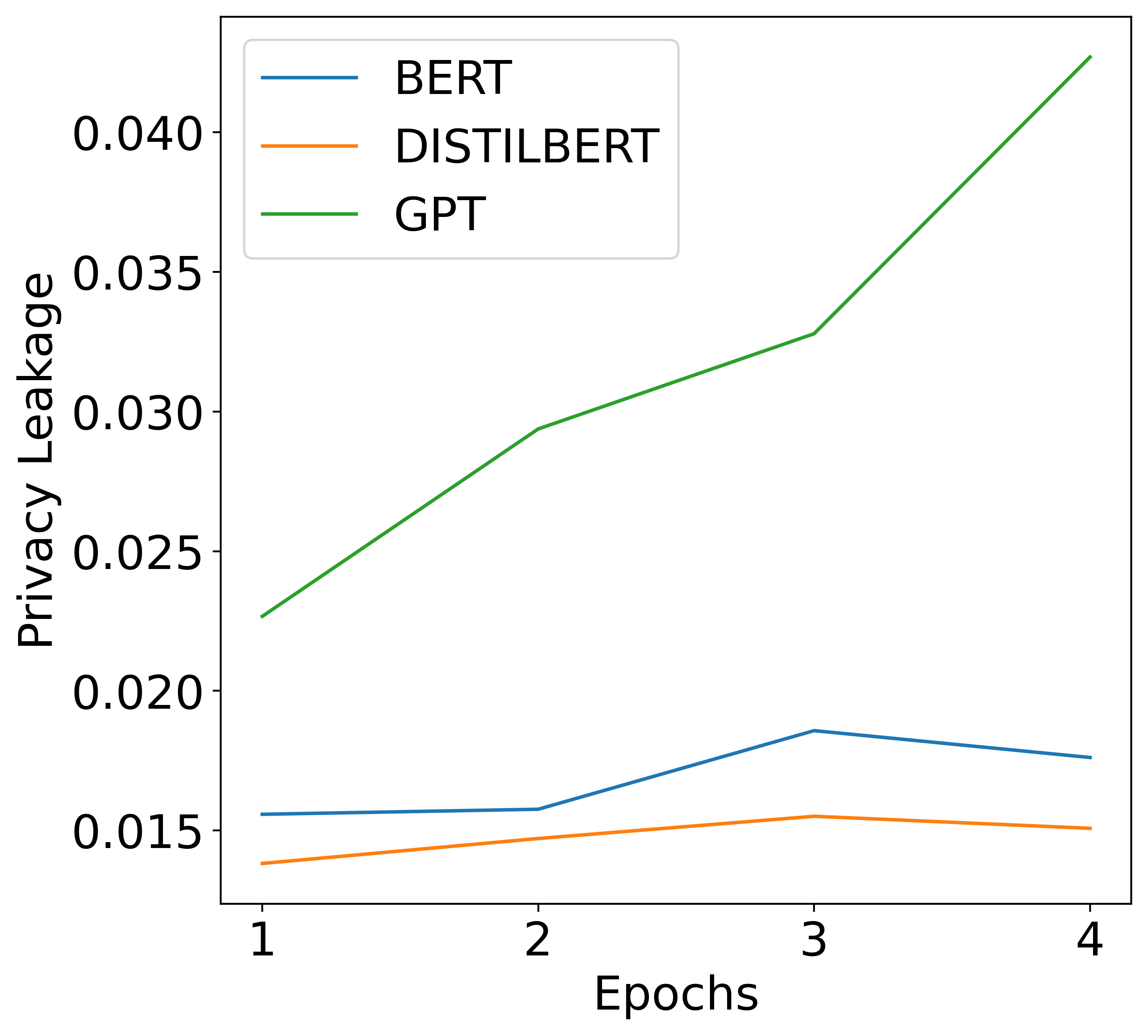}}
    \subfigure[]{\includegraphics[width=0.3\textwidth]{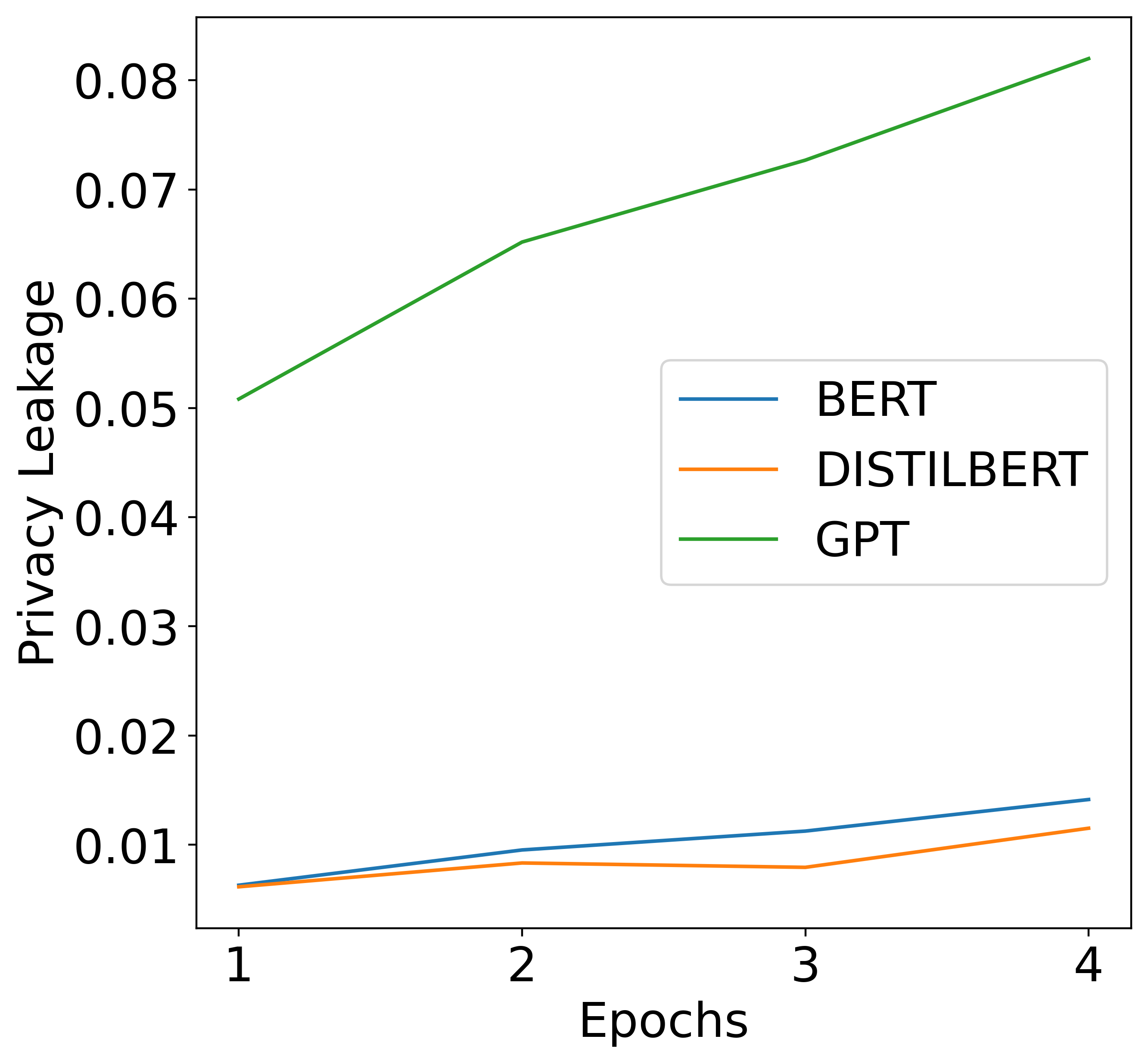}}
    \caption{Sample-level privacy leakage using black box and white-box membership inference attacks for non-DP CLM models. Plots show results on the MIMIC-III data. Plot (a) shows S-BBA attack leakage. Plot (b) and (c) show S-AWBA and S-GWBA white box attack leakages respectively. The \textsc{gpt2} model consistently has the highest leakage and Distil\textsc{bert} has the lowest.}
    \label{fig:mimic_sample_non_priv}
\end{figure*}

\section{Background}
\paragraph{Pre-trained Language Models:}
% \textcolor{red}{To do}
Language Models (LMs)\footnote{All useful notations are also provided together in Appendix~\ref{appendix:notations}.} such as \textsc{bert}, Ro\textsc{bert}a or \textsc{gpt2} are pretrained using large unsupervised text corpora such as WikiText that contains text from multiple domains. These pre-trained LMs are used for extracting contextual embeddings which are used for multiple downstream NLP tasks. 
For domain specific NLP tasks such as MedNLI or emrQA, the LMs are further fine-tuned on text from a particular domain such as clinical\textsc{bert} \cite{alsentzer2019publicly}, bio\textsc{bert} \cite{lee2020biobert} and sci\textsc{bert} \cite{beltagy2019scibert}.% and are released publicly.

\paragraph{Differential Privacy:}
\label{sec:DP}
$(\epsilon,\delta)-$DP \citep{dwork2014algorithmic} can be used to limit the privacy leakage of machine learning models trained through stochastic gradient descent(SGD). A randomized mechanism $\mathcal{M}:\mathcal{D}\rightarrow\mathcal{R}$ --- in our case, this refers to the SGD training algorithm ---  is $(\epsilon,\delta)$-DP compliant if for all $S \subset \mathcal{R}$ and $x,x' \in D$
\begin{equation}
    \mathrm{Pr}[\mathcal{M}(x)\in S]\leq e^\epsilon \mathrm{Pr}[\mathcal{M}(x') \in S] +\delta.
\label{eqn:diff_privacy}
\end{equation}

An ML model with $(\epsilon,\delta)$-DP compliant training limits the privacy leakage to a function of $\epsilon$ with failure probability $\delta$. An important property of $(\epsilon,\delta)$-DP systems is \textbf{\textit{Group Differential Privacy}}. Group DP provides $(\epsilon,\delta)$ values for $k$ correlated inputs in a dataset. In such a case the privacy degrades to $(k\epsilon, k e^{(k-1)\epsilon}\delta$) \citep{dwork2014algorithmic}. In private applications, ideally we would keep $\delta \leq 1/N$, for dataset size $N$ and $\epsilon<<1$.

\paragraph{Membership Inference:}
Machine learning model parameters 
can be 
probed to extract sensitive information about the training data samples \citep{homer2008resolving,yeom2018privacy}. \citet{shokri2017membership} introduce the membership inference attacks, which refer to a class of attack models that predict whether a given data sample was present in the training data for a trained model. The advantage of an attacker using membership inference attack depends on the attacker's ability to distinguish between the \textit{target} model's response to training and out-of-training data samples.  

\citet{shokri2017membership} proposed a shadow model based attack that can try to understand differences in model's response for training data samples vs out-of-training(OOT) data samples.  Computation and memory costs of shadow models scale with the number of model parameters and therefore, these attacks are not feasible for large CLM models such as \textsc{bert}. \citet{yeom2018privacy} propose black-box attacks that achieve privacy leakage similar to shadow model attacks. 

\section{Data}
\label{sec:data}
We use three US hospital datasets, namely MIMIC-III, UMM (UMass Memorial Health Care)and VHA (Veterans Health Administration) Hospitals for our study. These datasets are used to train our CLMs and evaluate their privacy leakage. MIMIC-III \citep{johnson2016mimic} is an publicly available dataset, while UMM and VHA  data sources are private. Our main experiments are conducted with MIMIC-III and UMM datasets. Due to \textit{in-house} computational constraints, we use VHA  dataset to only evaluate privacy leakage for one of our CLM models. We extract EHR repositories from all three data sources along with patient and admission level information. We select all available patient notes from MIMIC-III. We use a manageable subset of patients from UMM and VHA  repositories, detailed information provided in Appendix~\ref{appendix:training_data_details}.

In MIMIC-III, each patient's records are composed of a collection of EHRs of clinical visits or admissions. Each visit is composed of a group of EHRs that were recorded during that visit. Each EHR note is, in turn, a document that may be composed of multiple text samples. We use this hierarchical structure to study the effects of correlated samples on privacy leakage. In UMM and VHA , we could not use admission information. Therefore we treat each EHR record as an admission. Correlated samples can extend the notion of privacy to group-level privacy using the definition provided in Section \ref{sec:DP}. 
Dataset size statistics for all three datasets are provided in Table \ref{tab:datasets}.

\begin{figure}
    \centering
    \includegraphics[width=0.7\columnwidth]{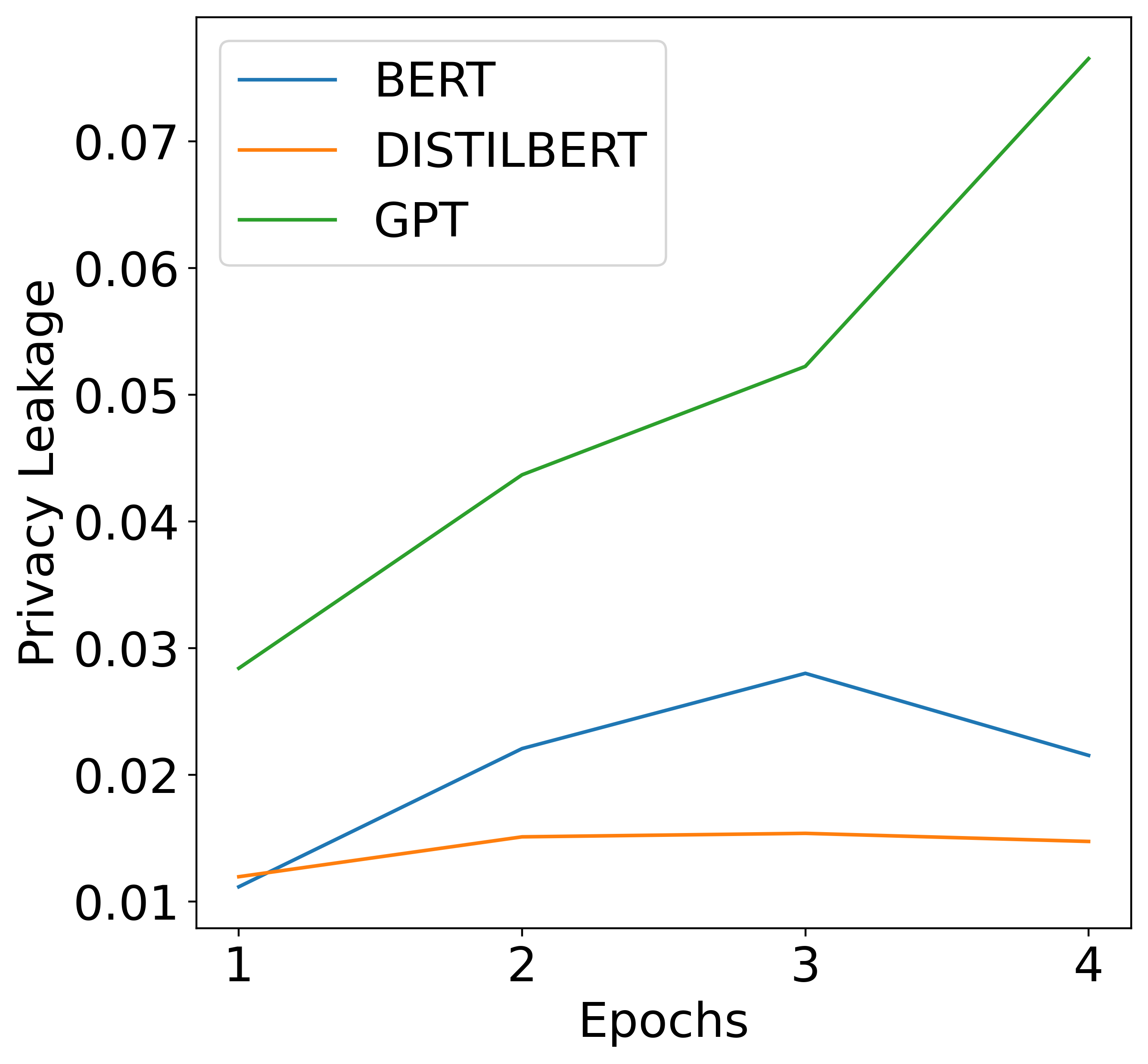}
    \caption{Admission-level membership attack (A-BBA) privacy leakage using MIMIC-III data for non-DP CLMs.}
    \label{fig:mimic_bb_group_non_priv}
\end{figure}

\section{Methods}
For our investigations, we use two main sets of methods. We use membership inference attacks to estimate empirical privacy leakage of a method. And we use differentially private stochastic gradient descent training (DP-SGD)\citep{abadi2016deep} to train DP clinical LM models.

\subsection{Membership Inference Attacks}
\label{sec:membership_inference}
Membership inference attacks are accomplished by exploiting the difference in a model's responses to training and OOT data samples. We use the \textit{advantage} of membership inference attacks to estimate the empirical privacy leakage. The experimental framework for obtaining privacy leakage using a membership inference adversary is defined by \citep{yeom2018privacy} as follows:

\paragraph{Membership Inference Experiment $\textrm{Exp}^M(\mathcal{A}, A, n, D)$:} Let $\mathcal{A}$ be an adversary, A be a learning algorithm, n be a positive integer, and $D$ be a distribution over data points $(x, y)$. The membership experiment in \citep{yeom2018privacy} proceeds as follows:
\begin{itemize}
    \itemsep0em
    \item Sample $S \in D_n$.
    \item Choose $b \leftarrow \{0, 1\}$ uniformly at random.
    \item Draw $z \in S$ if $b = 0$, or $z \in D$ if $b = 1$.
    \item $\textrm{Exp}^M(\mathcal{A}, A, n, D) = 1$ if $\mathcal{A}(z, A(S), n, D) = b$ and $0$ otherwise. $\mathcal{A}$ must output either $0$ or $1$.
\end{itemize}

\begin{figure}[t]
    \centering
    \includegraphics[width=0.7\columnwidth]{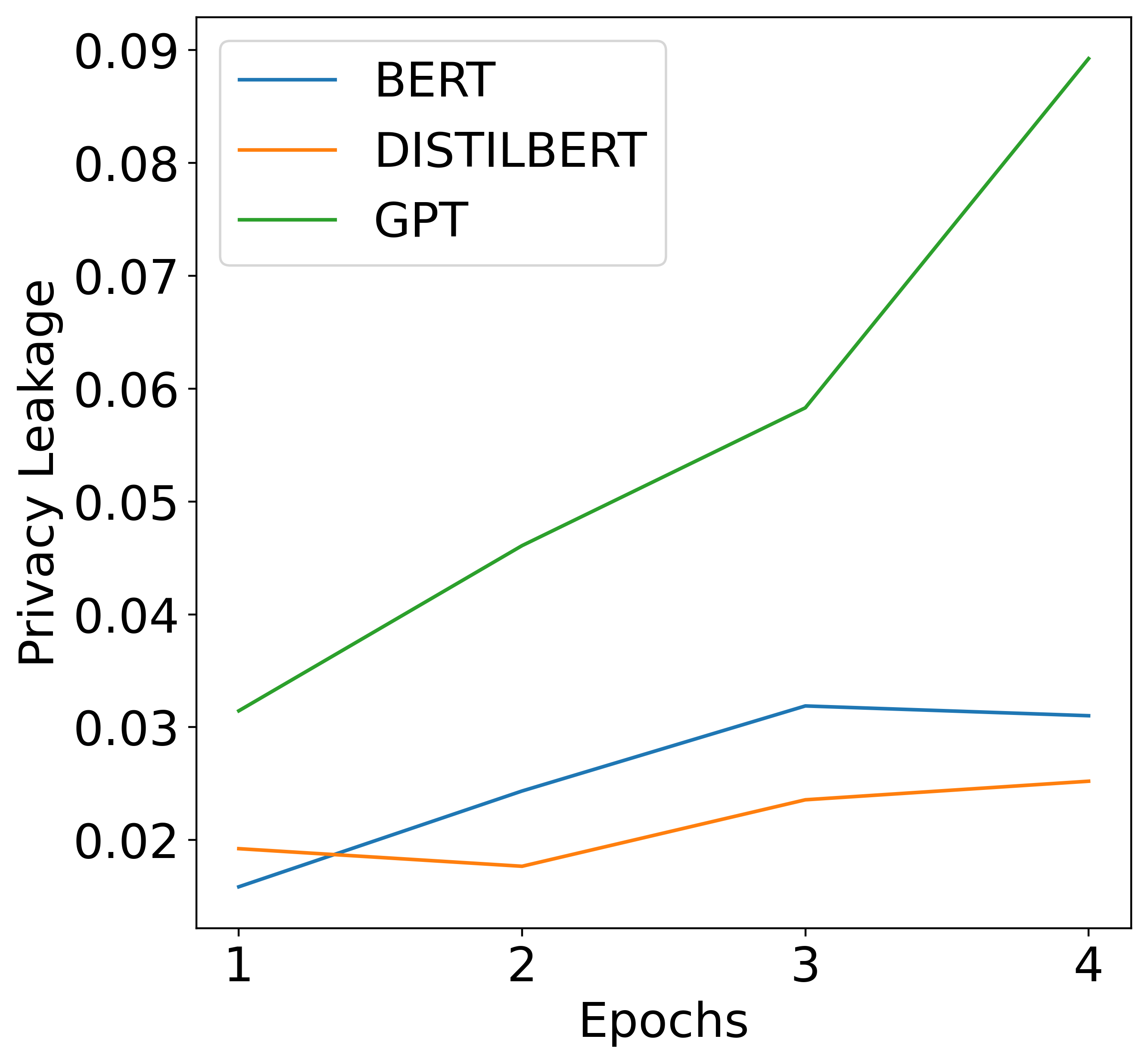}
    \caption{Patient-level membership attack (P-BBA) privacy leakage using MIMIC-III data for non-DP CLMs.}
    \label{fig:mimic_bb_group_non_priv_2}
\end{figure}

The membership advantage of the adversary $Adv^M$ or alternatively the empirical privacy leakage observed in the experiment is  
\begin{equation}
PL(\mathcal{A}) = Pr[\mathcal{A} = 0 | b = 0] -  Pr[\mathcal{A} = 0 | b = 1].
\label{eqn:PL}
\end{equation}
NLP models may be released with API or white-box access to the internal model. Therefore, we design both black-box and white-box adversaries for our study. 

\paragraph{Samples}
For membership inference experiments in CLMs, the sample $S$ consists of the input sentence and the constructed language modeling target. In case of an auto-regressive language model such as GPT-2, the input sequence is a fixed length token sequence and the target is the next token sequence. In case of MLM objective models like BERT, the input sequence is a randomly masked fixed length token sequence and the output is the relevant masked token values. It should be noted that masks are generated at random in MLM training procedures, so the masks generated for membership inference experiment will be different than those generated during training of the CLM. 

\subsubsection{Black-Box Adversary}

We adapt the black box attack defined by \citet{yeom2018privacy} (\textit{Adversary} 2). This attack model assumes access to the error probability density functions of the trained model, $f(\epsilon|b=0)$ and $f(\epsilon|b=1)$. 
\paragraph{Sample-level Attack}
For a sample $x$, the output of the adversary $\mathcal{A}(x)$ is $argmax_{b\in(0,1)}f(\epsilon(x)|b)$. In context of language models with non-zero error means, we can define $f(\epsilon|b)$ using error mean and unit variances for both b=1 and 0. However, we use a simpler scheme that only assumes access to the mean of the training error. In preliminary experiments this choice did not reduce the attack advantage. We use the mean of training error $\mu_{tr}$ as the threshold to predict $\mathcal{A}(x)$. Our sample level attack is very similar to threshold adversary proposed by \citet{yeom2018privacy}. A random sample $x$ is predicted to be a training data sample if $e(x)<\mu_{tr}$ else it is predicted to be an OOT data sample. We denote this sample-level black-box membership inference attack as \textit{\textbf{S-BBA}}. 
\paragraph{Group-level Attack}
To investigate group-level privacy leakages, we devise group-level membership inference attacks for our black-box access adversary. For group-level membership inference, we treat each group as a single data sample. To estimate the error of a group $g$ we take a mean of all samples in that group. We use the same condition and threshold ( $e(g)<\mu_{tr}$) as that in S-BBA for our group-level privacy attacks. The group-level attacks based on patient and admission level groups are referred as \textit{\textbf{P-BBA}} and \textit{\textbf{A-BBA}} respectively. 

\subsubsection{White-Box Adversary}
With white-box access to the model, we can use hidden layer outputs, gradients and other model specific information to identify differences in model's behavior between training and out-of-training data samples. Based on preliminary experiments, gradient and attention values proved to be more informative than hidden layer outputs.

\paragraph{Attention-based white-box attack (S-AWBA):} In this attack we use the self-attention outputs of CLMs as the input to the attack model. A k-head transformer layer with a sequence input length of n, produces $kn^2$ attention values. For a 12 layer LM such as \textsc{bert}, with 12 attention heads for each transformer layer, the number of attention outputs is $144n^2$. For an input sequence of 512 tokens, a model like \textsc{bert} will have around 37 million attention values. To reduce the size of attack model input, we instead use the ``concentration'' of the attention vector $a=\{a_i\}_{0}^{n}$ using 

\begin{equation}
C(a)=\sum_{i=0}^{n} a_i \log(a_i) .   
\end{equation}

$C(a)$ is higher when the attention is diffused across all token positions and lower when high attention values are focused on specific tokens. We obtain one $C(a)$ value for each token on each attention layer. To further reduce the size of the attention information, we use four estimates computed over the sentence length $\{C(a)_j\}_0^n$, namely mean, median, 5-percentile and 95-percentile values. Computation of $C(a)$ values and aggregation over the  sentence provides us with a compressed $144\times 4$ dimensional attention representation for any input sample. We use this attention representation as an input to the attention-based attack model. 

\paragraph{Gradient-based white-box attack(S-GWBA):} Gradient values have been previously \citep{nasr2019comprehensive} used in membership inference attack models against DNNs. Intuitively, the gradient values of a training sample's loss are expected to have lower absolute value than an OOT data sample's loss derivative. Membership inference attacks using gradient values from large LM models pose similar computational challenges as the attention-based attacks. An LM model, such as \textsc{bert} has 110 million parameter vectors. As a result, it produces 110 million partial derivatives. We aggregate the gradient values for each neural network layer by taking the squared-norm of all parameters in that layer. For a model like \textsc{bert}, this results in a $202$ dimensional aggregate. We use this gradient representation as an input for the attack model. 

\subsection{Differentially Private Model Training}
\label{sec:DPSGD}
We use differentially private SGD (DP-SGD) proposed by \citet{abadi2016deep} to train our $(\epsilon,\delta)-$DP models. DP-SGD uses per-sample gradient clipping to bound the effect of each sample and uses additive Gaussian noise to inject noise into the mini-batch gradients. In our experiments, we use the noise standard deviation (std) as a hyperparameter and denote it as $\sigma$. We use R\'enyi Differential Privacy (RDP) framework \citep{mironov2017renyi} to estimate the total privacy loss due to the DP-SGD training.

\begin{table}
\centering
\begin{tabular}{p{0.11\textwidth}|p{0.07\textwidth}|p{0.11\textwidth}|p{0.07\textwidth}}
\hline
Dataset & \#Token & \#Admission & \#Patient \\
\hline
MIMIC-III & 695M & 57k & 46k\\
UMM & 1.33B & 1184k & 58k\\
VHA  & 8.31B & 2019k & 35k\\
\hline
\end{tabular}
\caption{Number of tokens, admission and patients in the three datasets. The number of tokens are computed using \textsc{bert}-base-cased tokenizer. We do not have admissions information for UMM and 3, so we consider one EHR note per admission.}
\label{tab:datasets}
\end{table}

\begin{table*}
\centering
\begin{adjustbox}{width=0.97\textwidth}
\begin{tabular}{c|c|c|c|c|c|c|c}
\textbf{Model /}$\sigma$ & \textbf{2} & \textbf{1} & \textbf{0.1} & \textbf{1e-2} & \textbf{1e-3} & \textbf{1e-4} & \textbf{1e-20}\\
\hline
\textsc{gpt2} &\underline{0/.009}&\underline{0/.014} & 0/.014 & 0/.014 & 0/.017 & 4e-3/.016 & 8e-3/.018 \\

\textsc{BERT} & \underline{9e-2/.011} & \underline{2e-3/.011} & 3e-3/.010 & 1e-3/.011& 3e-4/.017 & 3e-3/.016 & 5e-3/.016\\
Distil\textsc{BERT} &  \underline{0/8e-3} & \underline{0/5e-2} & 0/9e-2 & 0/8e-3 & 1e-5/.010 & 1e-3/.010 & 0/.012\\

\hline
\end{tabular}
\end{adjustbox}
\caption{Empirical privacy leakage for DP models trained on MIMIC-III data using S-BBA and S-AWBA attacks. Black box S-BBA always leads to lower PL than white box S-AWBA attack. PL values reported in this table are the maximum PL values over all three epochs of these models. Underlined models have RDP computed $\epsilon < 1$. All other models have $\epsilon > 1$. RDP accounting results are in Appendix \ref{app:results}.}
\label{tab:pv_models}

\end{table*}

\section{Experiments and Results}
We run our experiments on non-DP and DP CLMs that are trained on EHR datasets described in Section \ref{sec:data}. DP and non-DP CLMs use SGD and DP-SGD training procedures. Training details are provided in Appendix \ref{app:experiments}. We use \textsc{bert}, Distil\textsc{bert} and \textsc{gpt2} base models to initialize the CLM training. These CLMs are used in the following sections to study the patterns of privacy leakage across datasets and models.

\subsection{What is the sample level empirical privacy leakage of CLM models?}
We examine privacy leakage for clinical-domain tuned CLMs using both white-box and black-box membership inference attacks on text snippets from electronic health records. MIMIC-III and UMM data are used for the main set of experiments. Due to \textit{in-house} computational constraints for VHA , we were unable to replicate all experiments on VHA  data. However we do provide black box inference attack based PL results for \textsc{bert} models in Appendix \ref{app:results}. The general trends discussed in the following sub-sections hold true also for the limited results from VHA . 

\paragraph{Experimental Design:} Formally, we estimate the leakage by calculating S-BBA, S-AWBA and S-GWBA adversary advantage for non-DP CLMs. To evaluate the privacy leakage using Equation \ref{eqn:PL}, we use the membership experiment defined in Section \ref{sec:membership_inference}. We draw a sample $z$ from the training or test data for $b$ = 0 or 1, respectively. 
Estimation of $\textrm{Pr}[\mathcal{A} = 0 | b = 0]$  and $\textrm{Pr}[\mathcal{A} = 0 | b = 1]$ use training and test data samples. White-box attacks use a multi-dimensional vector of gradient or attention aggregates. To estimate the difference in attention and gradient values for training and OOT data sample, we train a logistic regression model ($A$) on 20\% split of the train and test datasets. Recall from section \ref{sec:membership_inference} that $A$ is used to predict $b$ for a data sample $z$. The remaining 80\% data samples are used to estimate $\textrm{Pr}[\mathcal{A} = 0 | b = 0]$ and $\textrm{Pr}[\mathcal{A} = 0 | b = 1]$ for provided $\mathcal{A}$.

\paragraph{Results:} Privacy leakage plots using MIMIC-III for all non-DP CLMs are provided in Figures \ref{fig:mimic_sample_non_priv}. The privacy leakage of all models for all epochs remains lower than 10\%. Distil\textsc{bert} model that has the lowest number of parameters amongst all three models, on average has the lowest sample level privacy leakage. This behaviour is consistent across all attack methods, both black-box and white-box. A consistent trend in our results is that the white-box access attacks lead to higher PL values. This is expected. Gradient based white-box attack is the most effective attack. The leakiest model is \textsc{gpt2} with privacy leakages is as high as 7\%. On average, all PL values increase with the number of epochs.

\subsection{Can we train $(\epsilon,\delta$)-DP CLM models with improved clinical domain performance?}

Models with 7\% PL, as shown in the previous section, should be further studied before releasing them in the public domain. To this end, we use DP-SGD training methods to enforce privacy constraints. We then study the trade-off between the strictness of privacy constraints and model utility. 
\paragraph{Experimental Design:} We use the DP-SGD and RDP for DP model training and privacy accounting as defined in Section \ref{sec:DPSGD}. To limit the group based privacy degradation in $\epsilon$ calculation, we limit the number of samples from each patient to 50. We use Gaussian mechanism with $\sigma$ ranging from 2.0 to 1e-4 in the DP-SGD algorithm. We also use a negligible $\sigma$ of 1e-20 to ablate the effect of additive noise from DP-SGD training. 

\paragraph{Results:} We see a decline in empirical privacy leakage across all models for all attack methods when we apply a Gaussian noise within the range  \{1e-4, 2.0\}. An empirical privacy leakage for a sample-level  white-box attack (S-AWBA) and black box attack (S-BBA) are provided in Table \ref{tab:pv_models}. There is a significant decrease in empirical privacy leakage. Most models exhibit less than 1\% (often less than 0.1\%) privacy leakage. We do see increased privacy leakage through attention access in S-AWBA, however, these leakages are around 1-2\% and are negligible for Distil\textsc{bert} model. Empirical privacy leakages for all models-attack combinations are provided in the Appendix \ref{app:results} for MIMIC-III and UMM data. One consistent pattern across all results show that models with $\sigma \in \{2.0,1.0,0.1,0.01\}$ have negligible privacy leakages for all sample and group membership inference attacks.  For $\sigma \in \{0.001,0.0001,1e-20\}$ we see more than 1\% privacy leakages in S-AWBA and S-GWBA. The most successful attack on non-DP models is conducted using S-GWBA, providing PL as high as 2\% for \textsc{gpt2}. Group membership inference attacks are advantageous even for a DP model. Group PL for models are as high as 5\% for $\sigma = 1e-20$ (S-GWBA). In conclusion, group-level membership inference attacks and white box attacks are able to extract more than 1\% privacy leakages from our DP models for very low epsilon values. However, most models with  $\sigma \leq 1e-3$ show negligible privacy leakage. Results for other models on MIMIC-III and UMM datasets are available in Appendix \ref{app:results}. They show similar behaviour.
To understand the trade-off between $\sigma$ and model utility, we compute the LM loss and MedNLI performance for each non-DP model epoch. Model performances for first two epoch \textsc{bert} models (DP-SGD, epoch=1) on MIMIC-III data are provided in Table \ref{tab:mimic_bert_model_utility}.
All models with $\sigma \leq 0.01$ have improved MLM test loss as compared to \textsc{bert}-base-cased. $\sigma \in \{0.01,0.001\}$ provides MLM loss lower than the \textsc{bert}-base-cased model while ensuring that empirical PL from all attack methods are kept below 1\%. MedNLI experiments and results for MIMIC-III \textsc{bert} models are provided in Appendix \ref{app:experiments} and \ref{app:results} respectively. We observe that most MedNLI performances for DP \textsc{bert} models show improved model utility with accuracy between \textsc{bert}-base and Clinical\textsc{bert}. 

\subsection{Does group level information provide more advantage to the membership inference attacker ?}

\begin{table}
\centering
\begin{tabular}{p{0.18\textwidth}|p{0.11\textwidth}|p{0.09\textwidth}}
\hline
\textbf{Model / Epochs} & \textbf{1} & \textbf{2} \\
\hline
Non-DP \textsc{bert} & 0.59 & 0.57 \\
\hline
\hline
DP \textsc{bert}(2.0) & 6.15 & 7.02 \\
DP \textsc{bert}(1.0) & 5.65 &  6.76\\
DP \textsc{bert}(0.1) & 3.82 &  4.86\\
DP \textsc{bert}(1e-2) & 2.00 &  2.33\\
DP \textsc{bert}(1e-3) & 1.18 &  1.23\\
DP \textsc{bert}(1e-4) & 0.85 &  0.86\\
DP \textsc{bert}(1e-20) & 0.75 & 0.76\\

\hline
\end{tabular}
\caption{DP \textsc{bert} ($\sigma$) denotes \textsc{bert}-base-cased models trained on MIMIC-III using DP-SGD with noise std $\sigma$. MLM loss is obtained by validating the model on test split.Untrained \textsc{bert}-base-cased model's MLM loss on MIMIC-III test is \textbf{3.49.} DP models for $\sigma \leq 1e-3$ have improved model utility due to DP training. For non-DP models increasing epochs increases model utility, whereas for DP model utility decreases with increased epochs. }
\label{tab:mimic_bert_model_utility}
\end{table}

To study the phenomenon of privacy degradation due to correlated data samples, we use admission and patient level membership inference attacks as defined in Section \ref{sec:membership_inference}.

\paragraph{Experimental Details:} We use the patient and admission ids provided in MIMIC-III dataset to construct the groups. For UMM data, we did not have access to admission level information, so each EHR is treated as one admission. The patient ids are used to create patient groups. The threshold used for the A-BBA and P-BBA attack adversaries are the same as that computed for S-BBA attack.

\paragraph{Results:} 
Group level privacy attacks for MIMIC-III and UMM are provided in Figures \ref{fig:mimic_bb_group_non_priv} and \ref{fig:mimic_bb_group_non_priv_2}. Comparing these plots with Figure \ref{fig:mimic_sample_non_priv} we observe an increase in the empirical privacy leakage for all models when group-level averages are used. The performance of larger group samples (patient-level aggregation) is higher than smaller group samples (admission or EHR level aggregation). This gap is more pronounced for UMM (figure in Appendix) where the EHR level group is much smaller than the patient-level group.

\section{Discussion}
\paragraph{Leakage in Non-DP models:}
Our results are consistent with existing efforts connecting over-fitting and privacy leakage \citep{yeom2018privacy}. Large LM models pretrained with dataset sizes in millions tend to have lower generalization gaps compared to supervised deep neural net models trained on more limited, expensive labelled data. Therefore we see a maximum sample level privacy leakage of around 7\% for non-DP CLMs. Additionally, losses like MLM use a random mask to produce a self-supervised sample. This random mask adds an additional layer of obfuscation, thereby reducing the empirical privacy leakage.
\textsc{gpt2} model, which uses an auto-regressive loss has higher privacy leakage than an equal sized MLM model. Distil\textsc{bert} LM model that has the smallest network size, shows the lowest privacy leakage as expected.

\paragraph{Trade-offs with privacy budgets:}
For theoretical privacy budgets computed using RDP accountant, only DP-SGD models with $\sigma \in {1,2}$ provide an $\epsilon<1$.  All other models with $\sigma < 1.0$ have $\epsilon$ values greater than 1, with the lower values like $\sigma=1e-3$ having $\epsilon$ values as high as $10^4$. $\epsilon$ computed for all models are provided in Appendix \ref{app:results}. The privacy leakage predicted by RDP is a vast overestimate of our empirical PL values. All models with $\sigma>1e-3$ show negligible empirical privacy leakage for both black-box and white-box attacks. Privacy budgets computed through DP accountants are vastly conservative and in the current form are not useful for deep neural net based NLP models. Based on our experiments, we find that keeping $\sigma$ higher than 1e-3 may prevent non-trivial privacy leakages from white-box membership inference and black-box group membership inference attacks. Conversely we see high model utility (close to non-DP model utility) for low  $\sigma$ values (Table \ref{tab:mimic_bert_model_utility}). According to group DP, for a group size of k, $\epsilon$ values should degrade to $50\times \epsilon$ and $\delta$ should degrade even more. While group level membership inference attacks increase the privacy leakage, they do not show such significant privacy degradation. This may be partially due to the nature of the EHR dataset, that has large text spans of standard instructions and copy-pasted text. This text results in several nearly-identical data samples across patients in both training and OOT data sets. It is also possible that our group-level attack is too simplistic, and more sophisticated methods are needed. 
\paragraph{Practical Attack Scenarios:}
As discussed in Section \ref{sec:intro}, membership inference attacks against an NLP system may not be applicable in a practical scenario due to the high dimensional input space and the lower rank semantic space of sentences. Our attacks can be used practically when partial information about the patient, such as a snippet of their EHR, is known to the attacker. Nevertheless, membership inference attacks serve as a good standardized framework for comparing relative privacy leakage of models and study the effects of enforcing privacy constraints on them.

\paragraph{Privacy for Rare Disease Patients:}
To understand if patients the behaviour of privacy leakage for outlier patients (or those with rarer disease profiles), we examine the black-box patient membership inference attacks in more detail for MIMIC-III trained CLMs. We group patients into buckets based on the rarity of their disease profiles. We use the coded ICD (International Classification of Diseases) entries for each patient to estimate the probability of that patient's disease profile. Details for the probability estimation are provided in Appendix \ref{app:experiments}.
 
We divide the patients into 100 buckets based on log-normalized ranges of disease profile probability. We discard buckets with less than 10 patients. We calculate privacy leakage using Equation \ref{eqn:PL} for each bucket individually. We use the same threshold used in S-BBA attack. We observe that for most non-DP models with epoch $> 1$, \textit{\textbf{bucket's average probability is inversely correlated to privacy leakage}}. Patients with rarer disease profiles tend to exhibit higher privacy leakage. The relevant graphs are provided in Appendix \ref{app:results}.

\section{Related Work}
% \textcolor{red}{To Do}
Pre-trained Language Models are extensively used in current state-of-the-art pipelines for different NLP tasks. These models are generally pre-trained using large unsupervised text from multiple domains \cite{devlin2018bert, liu2019roberta, radford2019language}. For domain-specific tasks such as MedNLI or emrQA, these LMs are further fine-tuned on domain-specific data which could be private and not publicly available \cite{alsentzer2019publicly, lee2020biobert, beltagy2019scibert}. 

The language models when fine-tuned on private data and released publicly are prone to privacy attacks and can leak individual training examples \cite{carlini2020extracting}. This poses even higher risks when the models are trained on private clinical data as it may lead to leaking sensitive patient information. \textit{Membership inference attack} \cite{hisamoto2020membership, nasr2019comprehensive} is the most common privacy attack, where the adversary tries to predict whether a particular example was used for training the model or not. \textit{Model inversion attack} reconstruct representative views of the training examples. \cite{carlini2020extracting} showed that large LMs such as \textsc{GPT}-2 can memorize sentences and using extraction attack were able to extract verbatim sequences from the training set including identifiable information (public) such as names, phone numbers and email addresses. To counter these attacks, differential private training techniques \cite{abadi2016deep, mcmahan2018learning} are used to train deep learning models. These differentially private training techniques can lead to a reduction in model accuracy \cite{jayaraman2019evaluating} and increase the pre-training time, which is quite significant for large LMs.

\section{Conclusion and Future Work}
Our experiments show that large LM models exhibit lower privacy leakages compared to supervised DNNs studied by \citet{jayaraman2019evaluating, shokri2017membership, yeom2018privacy}.

\textsc{bert} and \textsc{gpt2} do still exhibit non-trivial privacy leakages of up to 7\% for white-box attacks. In private datasets like EHR repositories, this suggests potential susceptibility to training data extraction attacks. While the exact privacy leakage requirements are subjective to the application, we should strive for lower than 1\% PL values in clinical domain. Our experiments with DP-SGD training show that when used with low $\sigma$ values, it can reduce empirical privacy leakages to less than 1\% while maintaining improvements in model utility during training. We also show that patients with rarer diseases may exhibit higher privacy leakages. Our work represents the first standardized comparison of privacy leakages in commonly used LM architectures. 

\paragraph{Future Work:} This work is the first step in studying privacy properties of CLMs, and has several future directions. A major research direction is to propose better model-agnostic and model-specific membership inference and training data extraction attacks. Our evaluation framework, along with a publicly available MIMIC-III is well suited to provide a testing framework for privacy attacks. Our future work involves establishing such a benchmark that provides the DP and non-DP models to researchers who already have access to MIMIC-III. An automated testing framework that computes patient-level privacy leakage can be established using our defined data splits. Another research direction is the use of data-dependent DP methods such as PATE \cite{papernot2018scalable}, which may be able to provide better $(\epsilon,\delta)$ values. Use of PATE for training large LMs is prohibited by the high computational cost of multiple-teacher training in PATE, and further study to reduce the computational complexity of such methods is required.  

\bibliography{acl2021}
\bibliographystyle{acl_natbib}

\clearpage
\appendix
\section{Experimental Details}
\label{app:experiments}
\subsection{Training Details}
We use the same procedure as used by clinical\textsc{bert} and bio\textsc{bert} to train our non-DP models. We split the total number of patients for each data source into a 70-30 train-test split. The dataset statistics for each data source is presented in Table \ref{tab:datasets}. \textsc{bert}, Distil\textsc{bert} and \textsc{gpt2} are trained on all notes from training split using non-DP SGD based training. These models are referred to as non-DP models. The learning rate of $1e-5$ is used and the models are trained for 4 epochs. 

We use DP-SGD with a gradient clipping threshold of 1 and varying $\sigma$ values to produce $(\epsilon,\delta)-$DP models with varying $\epsilon$ budgets. Allowed failure probability $\delta$ is kept constant at $1e-6$. To limit the privacy degradation due to group-level privacy losses, we limit each patient's contribution to the training dataset to 50 randomly selected samples. To manage additional computational complexity introduced by per sample gradient clipping, we run DP-SGD training for 3 epochs instead of 4 in the case of non-DP models. We refer to these models as DP models in our paper.

\subsection{Estimation of Patient Disease Profile Probability}
\label{app:disease_profile_calculation}
ICD coding uses a standardized vocabulary (ICD9 or ICD10) to document the relevant disease for a patient. For an admission, a patient may be coded with multiple ICD codes depending on the presentation of their symptoms and the course of diagnosis. To simplify our analysis, we assume that each ICD code is attributed independently. As a result the probability of observing a patient with $\textbf{c}=\{c_1,...c_k\}$ codes is
\begin{equation}
    p(\textbf{c})=\prod_{\textbf{c}} p(c_i) \prod_{\mathcal{C} \setminus \textbf{c}} (1-p(c_i)).
\end{equation} 

Here $\mathcal{C}$ is the set of all ICD codes and $\textbf{c}$ is the set of ICD codes for a patient. Since we assume that each ICD code is attributed independently, maximum likelihood estimate of any code $c'$ is 
\begin{equation}
    p(c')=\frac{\#\,of\,patients\,with\,code\,c'}{\#\,of\,total\,patients}.
\end{equation}

The same procedure can be followed for admission or note level analysis. 

\subsection{MedNLI Experiment Details}

All the model training details for MedNLI experiments are provided in Table~\ref{tab:mnli_datapoints}. All the models were trained over 8 seeds and their final performance was calculated by averaging over the top three performance values. MedNLI is a classification dataset and hence accuracy is used as an evaluation metric for all models.
\begin{table}[!htbp]
\centering
\begin{tabular}{lr}
\toprule
\multicolumn{2}{l}{\textbf{MedNLI Dataset}} \\ \midrule
Train datapoints  & $12,626$ \\ 
Test datapoints & $1,421$ \\ 
Learning Rate & $5e-5$\\
Batch Size & $12$ \\
Max Seq. Len. & $200$\\
Epochs & $3$\\ \bottomrule
\end{tabular}
\caption{Training details for MedNLI experiments.}
\label{tab:mnli_datapoints}
\end{table}

\subsection{Training Data Details}
\label{appendix:training_data_details}
MIMIC-III \cite{johnson2016mimic} is a publicly available critical care database made available via \url{https://mimic.physionet.org/}. 

Hospital 2 data is collected by selecting patients who are suffering from any form of cardiovascular diseases or patients who are suffering from any cancerous diseases. The patients were selected with the help of ICD9 and ICD10 codes mentioned in their structured medical data. 
The ICD codes for cardiovascular diseases were used according to \url{https://www.questdiagnostics.com/dms/Documents/Other/PDF_MI4632_ICD_9-10_Codes_for_Cardio_38365_062915.pdf}. The ICD codes for cancerous diseases were used according to \url{https://seer.cancer.gov/tools/conversion/2014/ICD9CM_to_ICD10CM_2014CF.pdf}

VHA  data was collected by randomly selecting patients who had hospital visits during the year 2020. All 2020 EHRs for the selected patients were extracted to produce the dataset.

\section{Detailed Results and Plots}
\label{app:results}

\subsection{RDP accounting for DP-SGD models}
We use RDP \citep{mironov2017renyi} accountant \footnote{https://github.com/tensorflow/privacy} to calculate $\epsilon$ values for $\delta=1e-6$. For MIMIC-III DP \textsc{bert} models (epoch 1) with $\sigma$ values of 2.0, 1.0, 0.1, the $\epsilon$ values are 0.223, 0.626, 403. For epoch 3 the $\epsilon$ values are 0.223, 0.628, 733. Models with $\sigma < 0.1$ have very large $\epsilon$ values. Since RDP accounting is independent of model accounting, $\epsilon$ values for Distil\textsc{bert} and \textsc{gpt2} models show similar behavior. Group DP $\epsilon$ values can be obtained by multiplying $\epsilon$ with group size ($k=50$ for our experiments). 

The RDP calculated $\epsilon$ values for UMM \textsc{bert} models (epoch 1) with $\sigma$ values of 2.0, 1.0, 0.1, are 0.219, 0.553, 391. For epoch 3 the $\epsilon$ values are 0.220,0.553, 619. Similar to MIMIC-III $\epsilon$ values, models with $\sigma < 0.1$ have very large $\epsilon$ values.

\subsection{Detailed Results for MIMIC-III Dataset}
In this section we provide privacy leakage, and model utility results for non-DP and DP models trained on MIMIC-III dataset. 

\subsubsection{Privacy Leakage Results}

\begin{figure*}
    \centering
    \includegraphics[width=1.0\columnwidth]{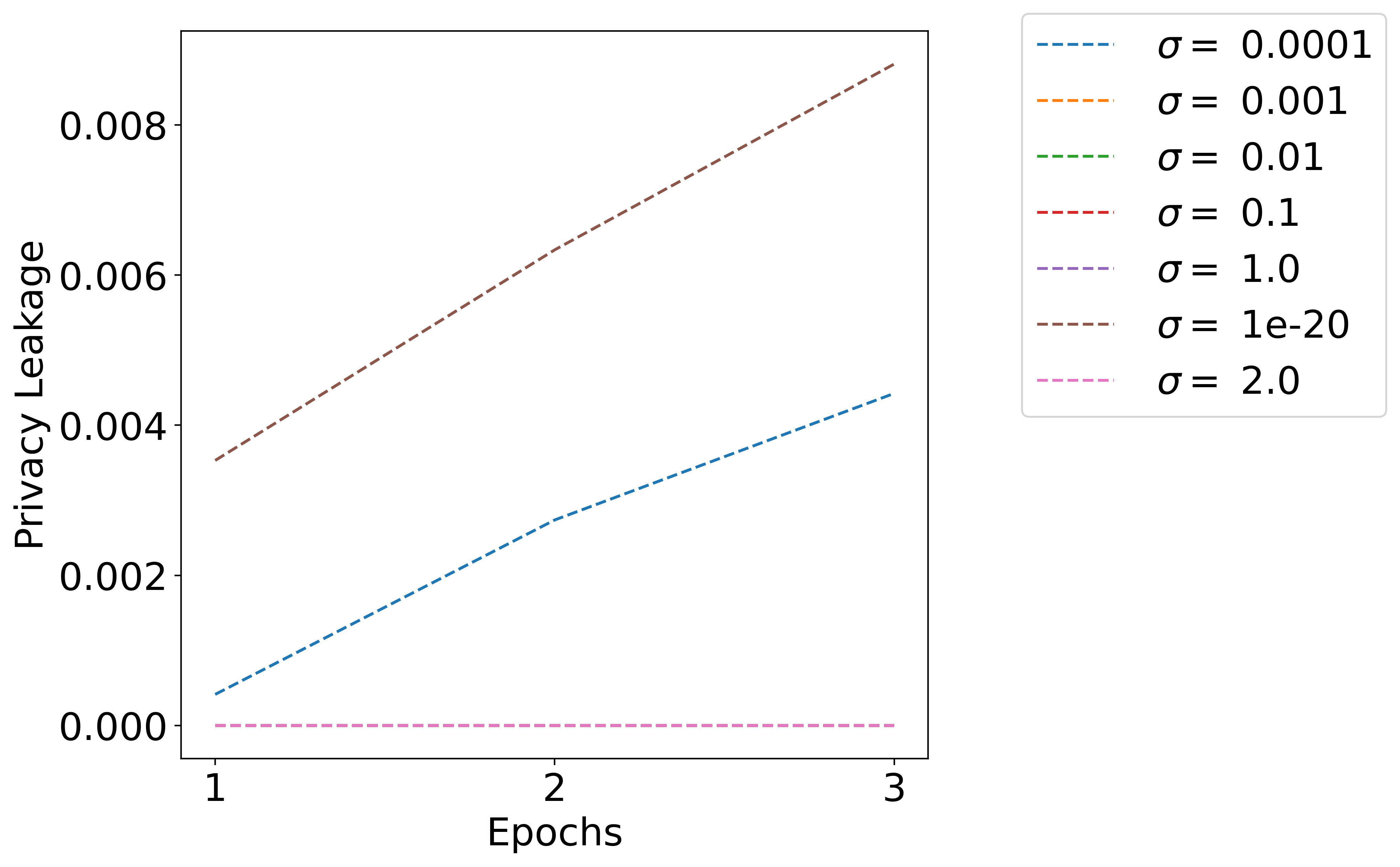}
    \caption{Privacy leakage for \textsc{gpt2} models with varying $\sigma$ values using MIMIC-III data. Leakage obtained using sample level black box attack (S-BBA). Gradient clip value is 1, group samples are limited to 50.}
    \label{fig:mimic_bb_sample_gpt2_priv}
\end{figure*}

\begin{figure*}
    \centering
    \includegraphics[width=1.0\columnwidth]{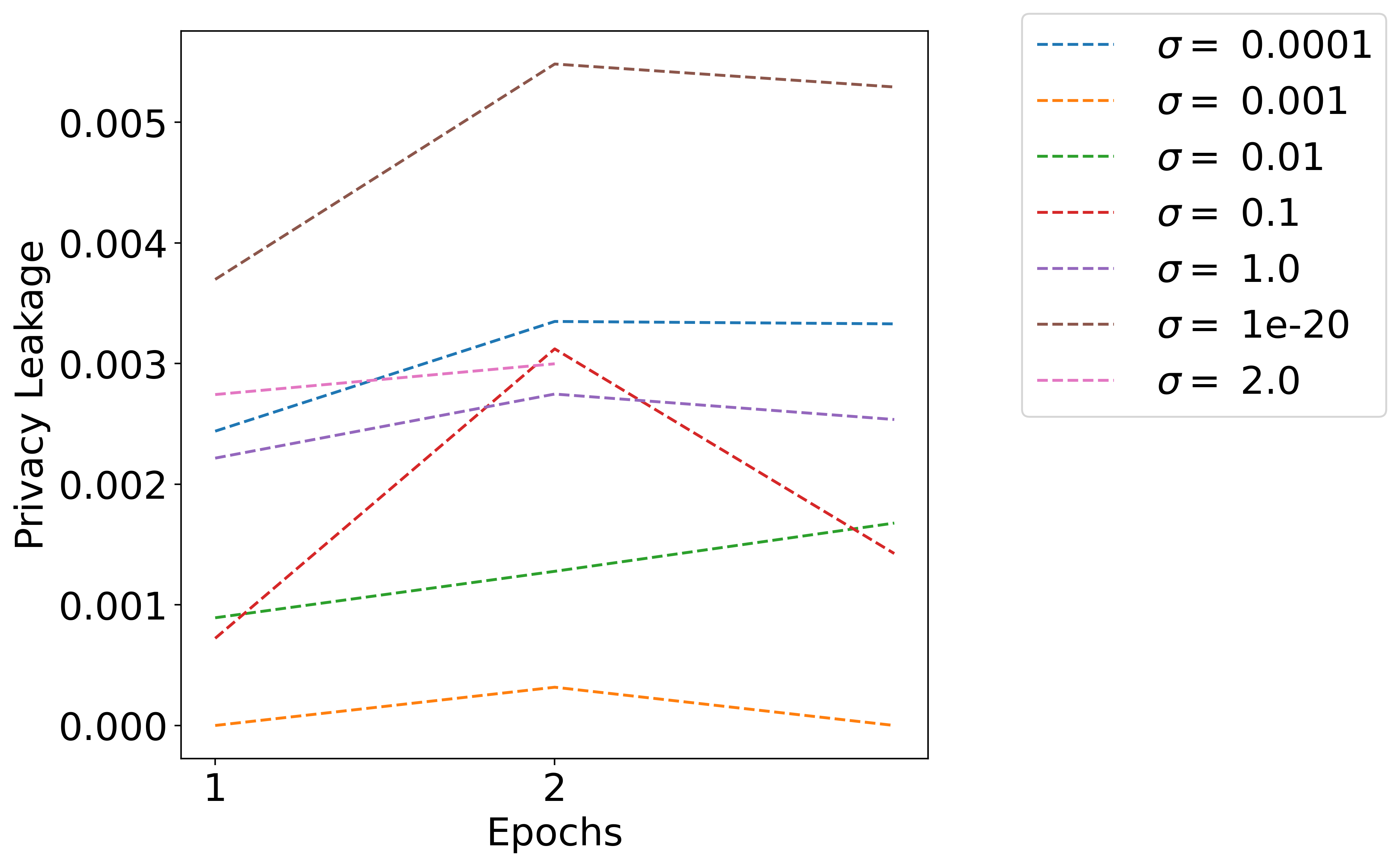}
    \caption{Privacy leakage for \textsc{bert} models with varying $\sigma$ values using MIMIC-III data. Leakage obtained using sample level black box attack (S-BBA). Gradient clip value is 1, group samples are limited to 50.}
    \label{fig:mimic_bb_sample_bert_priv}
\end{figure*}

\begin{figure*}
    \centering
    \includegraphics[width=0.9\columnwidth]{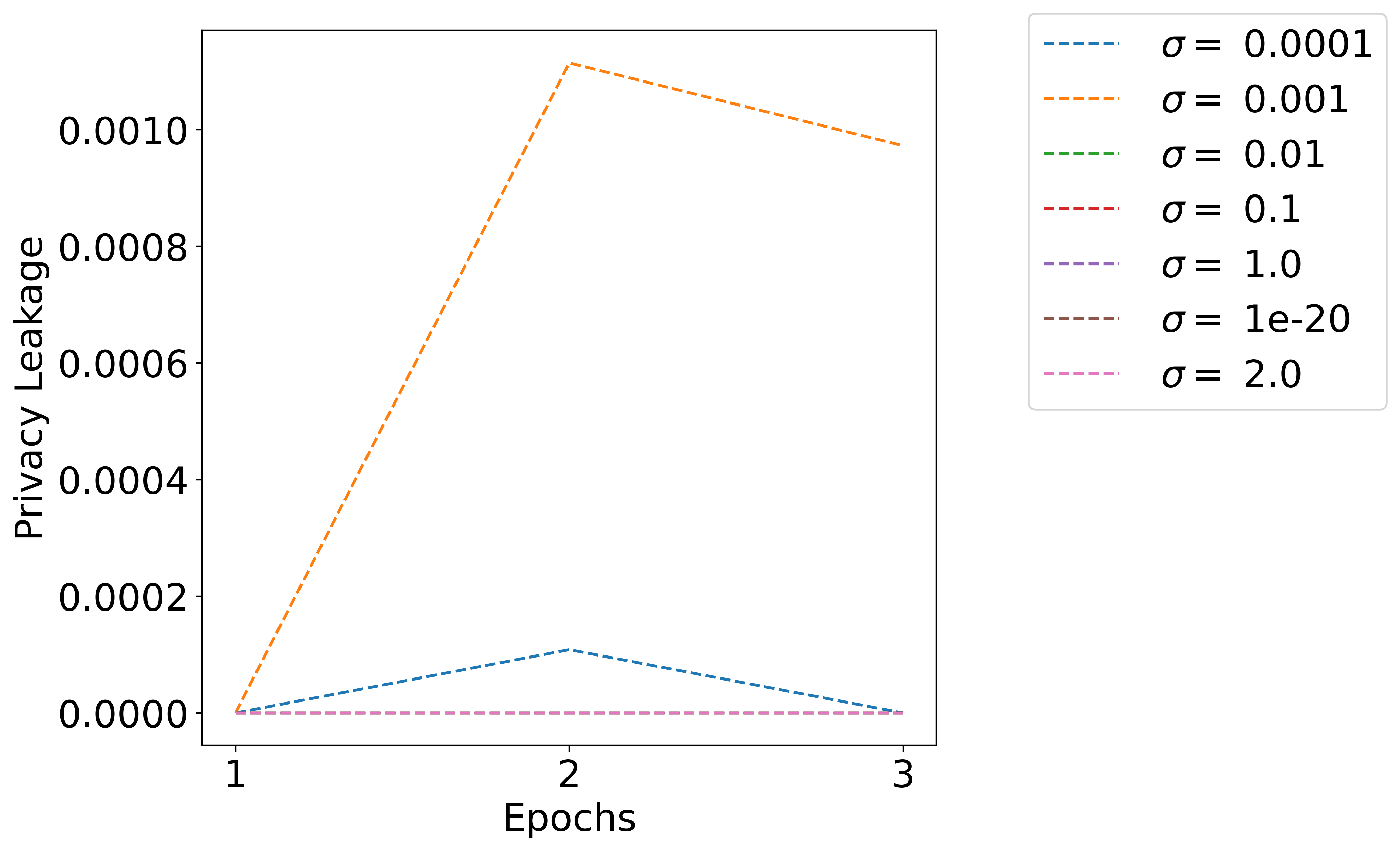}
    \caption{Privacy leakage for Distil\textsc{bert} models with varying $\sigma$ values using MIMIC-III data. Leakage obtained using sample level black box attack (S-BBA). Gradient clip value is 1, group samples are limited to 50.}
    \label{fig:mimic_bb_sample_distilbert_priv}
\end{figure*}

Figures \ref{fig:mimic_bb_sample_bert_priv}, \ref{fig:mimic_bb_sample_distilbert_priv} and \ref{fig:mimic_bb_sample_gpt2_priv} show sample level black box (S-BBA) attack privacy leakage for \textsc{bert}, Distil\textsc{bert} and \textsc{gpt2} models. 

\begin{figure*}
    \centering
    \includegraphics[width=0.9\columnwidth]{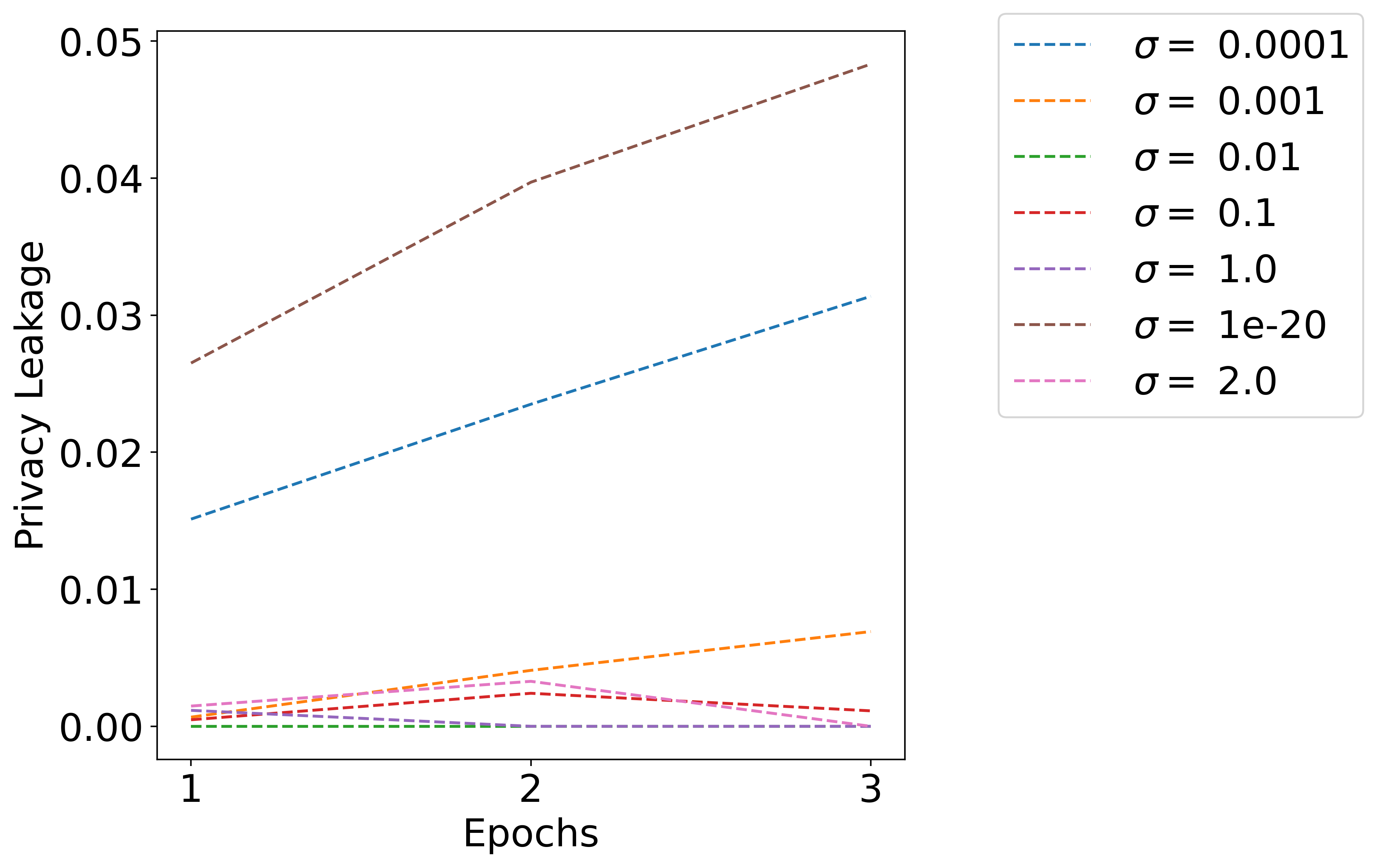}
    \caption{Group Privacy leakage for \textsc{gpt2} models with varying $\sigma$ values using MIMIC-III data. Leakage obtained using admission level black box attack (A-BBA). Gradient clip value is 1, group samples are limited to 50.}
    \label{fig:mimic_bb_hadm_gpt2_priv}
\end{figure*}

\begin{figure*}
    \centering
    \includegraphics[width=0.9\columnwidth]{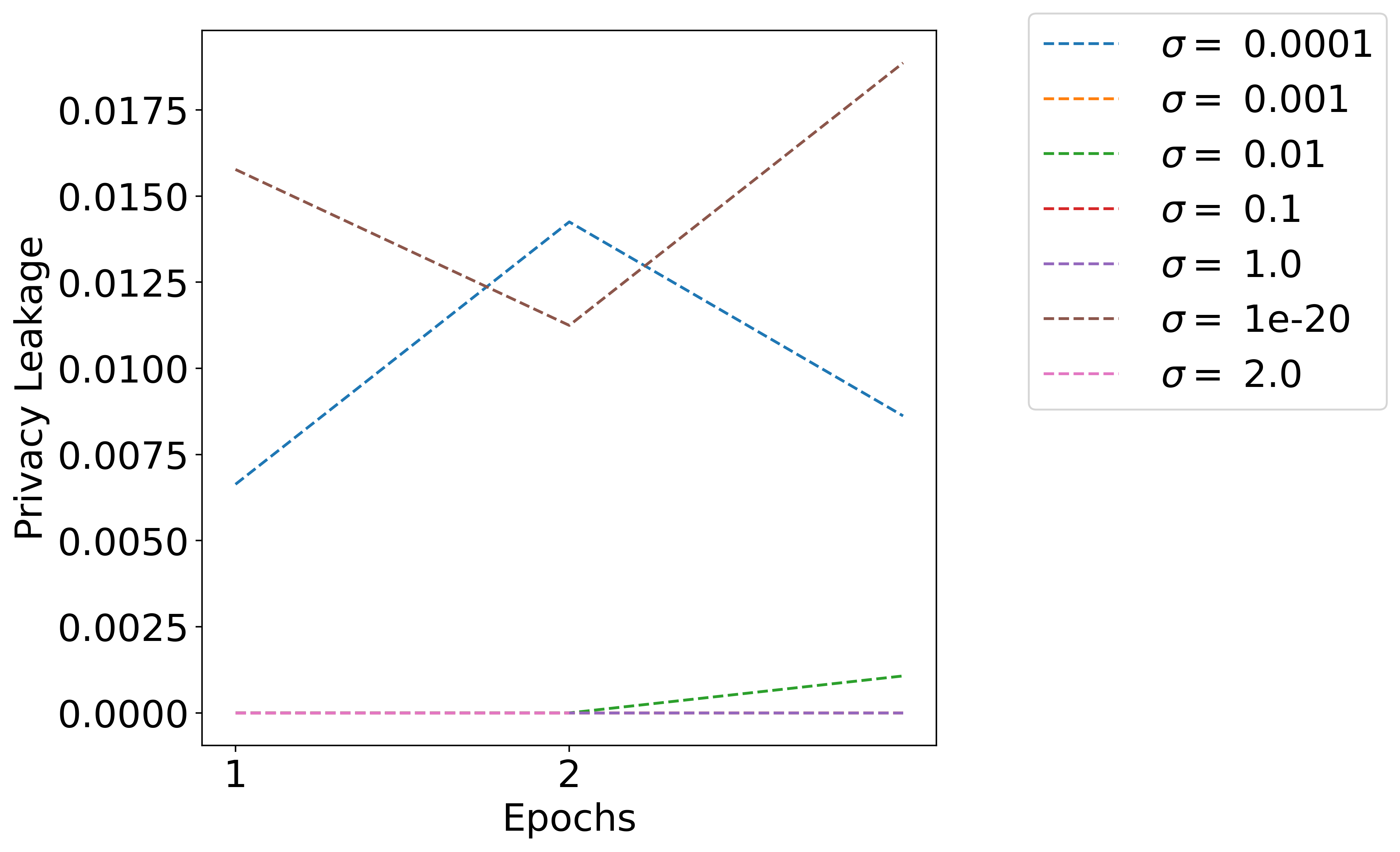}
    \caption{Group privacy leakage for \textsc{bert} models with varying $\sigma$ values using MIMIC-III data. Leakage obtained using admission level black box attack (A-BBA). Gradient clip value is 1, group samples are limited to 50.}
    \label{fig:mimic_bb_hadm_bert_priv}
\end{figure*}

\begin{figure*}
    \centering
    \includegraphics[width=0.9\columnwidth]{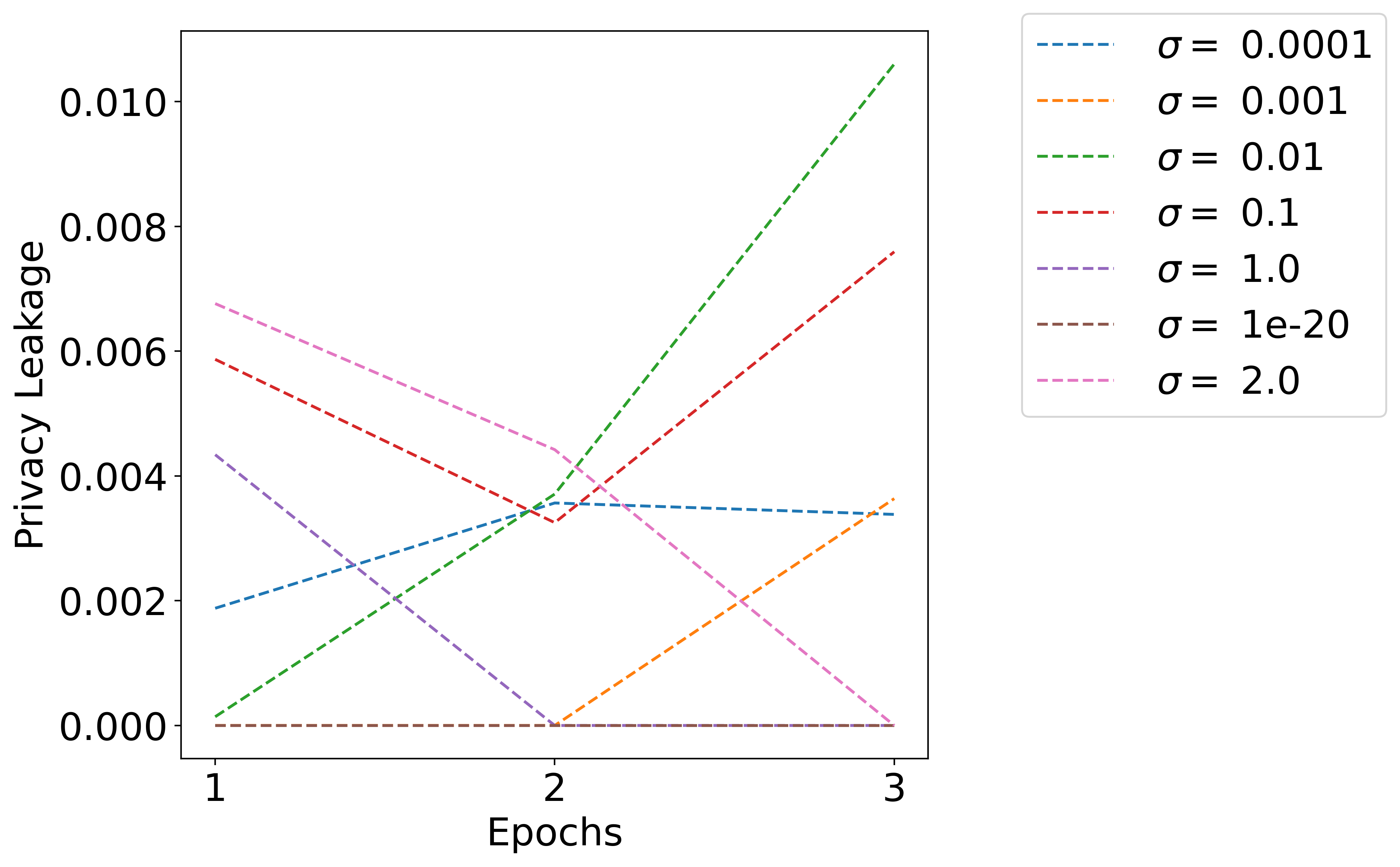}
    \caption{Group privacy leakage for Distil\textsc{bert} models with varying $\sigma$ values using MIMIC-III data. Leakage obtained using admission level black box attack (A-BBA). Gradient clip value is 1, group samples are limited to 50.}
    \label{fig:mimic_bb_hadm_distilbert_priv}
\end{figure*}

Figures \ref{fig:mimic_bb_hadm_bert_priv}, \ref{fig:mimic_bb_hadm_distilbert_priv} and \ref{fig:mimic_bb_hadm_gpt2_priv} show admission level black box (A-BBA) attack privacy leakage for \textsc{bert}, Distil\textsc{bert} and \textsc{gpt2} models. 

\begin{figure*}
    \centering
    \includegraphics[width=0.9\columnwidth]{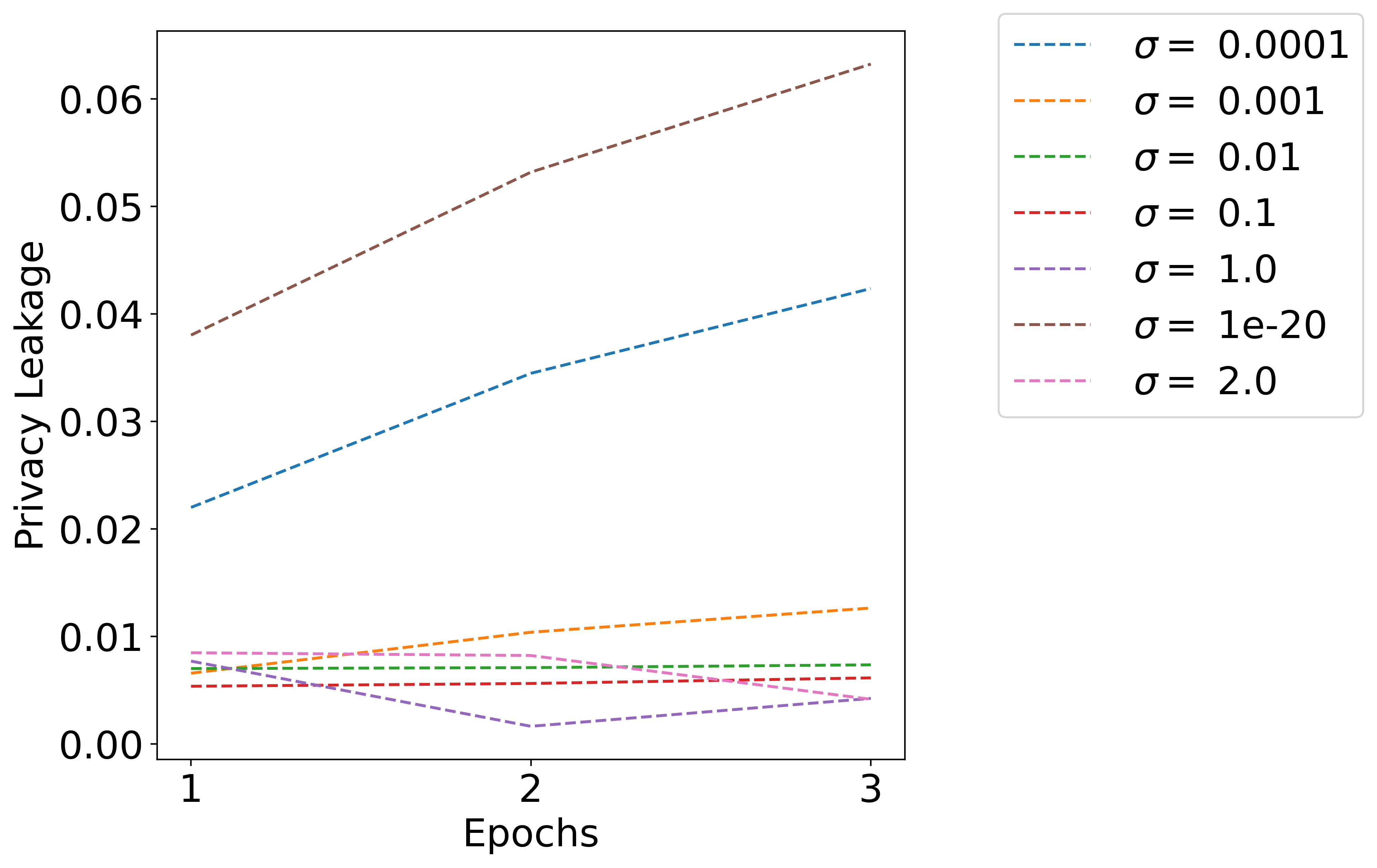}
    \caption{Group privacy leakage for \textsc{gpt2} models with varying $\sigma$ values using MIMIC-III data. Leakage obtained using patient level black box attack (P-BBA). Gradient clip value is 1, group samples are limited to 50.}
    \label{fig:mimic_bb_patient_gpt2_priv}
\end{figure*}

\begin{figure*}
    \centering
    \includegraphics[width=0.9\columnwidth]{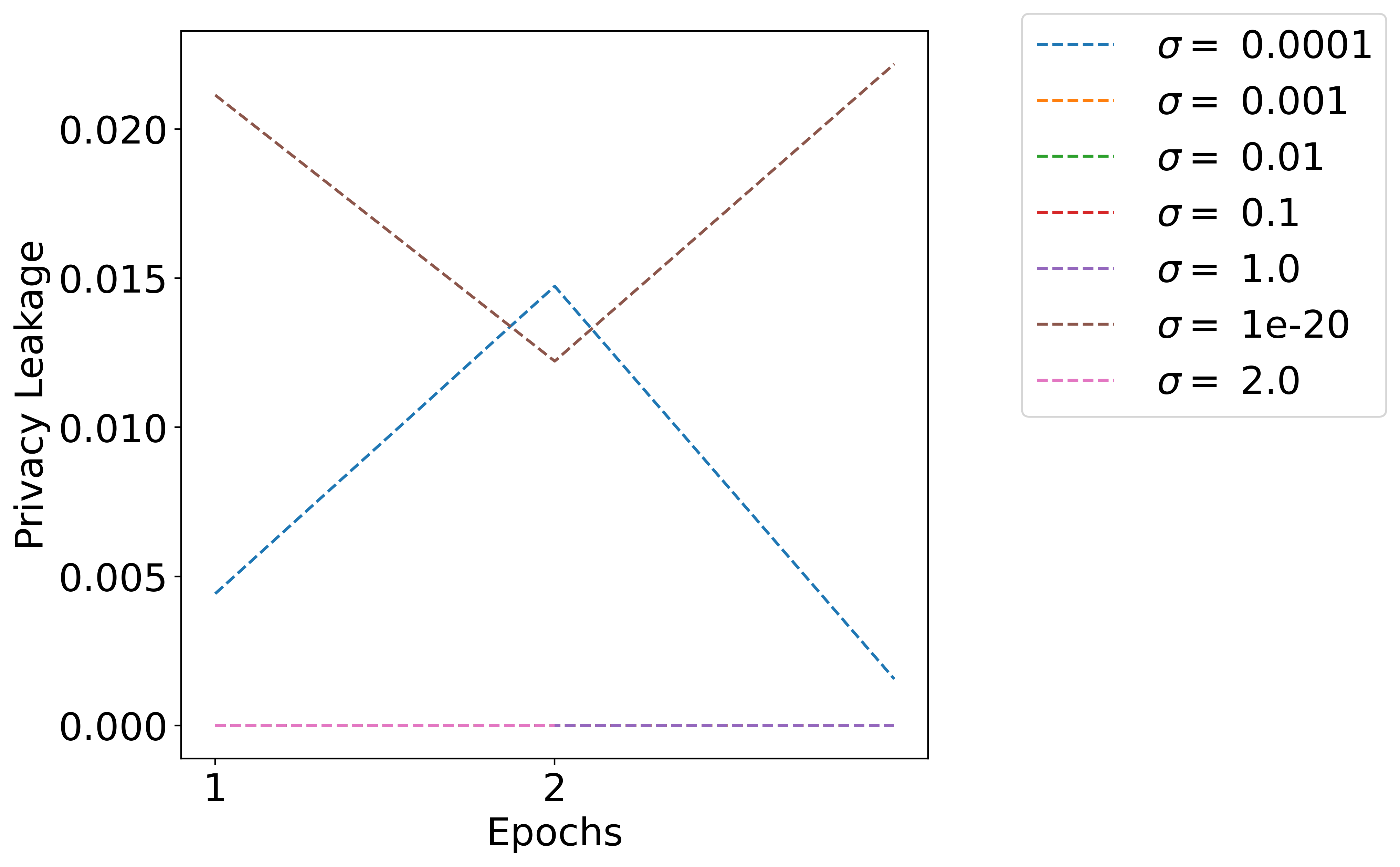}
    \caption{Group privacy leakage for \textsc{bert} models with varying $\sigma$ values using MIMIC-III data. Leakage obtained using patient level black box attack (P-BBA). Gradient clip value is 1, group samples are limited to 50.}
    \label{fig:mimic_bb_patient_bert_priv}
\end{figure*}

\begin{figure*}
    \centering
    \includegraphics[width=0.9\columnwidth]{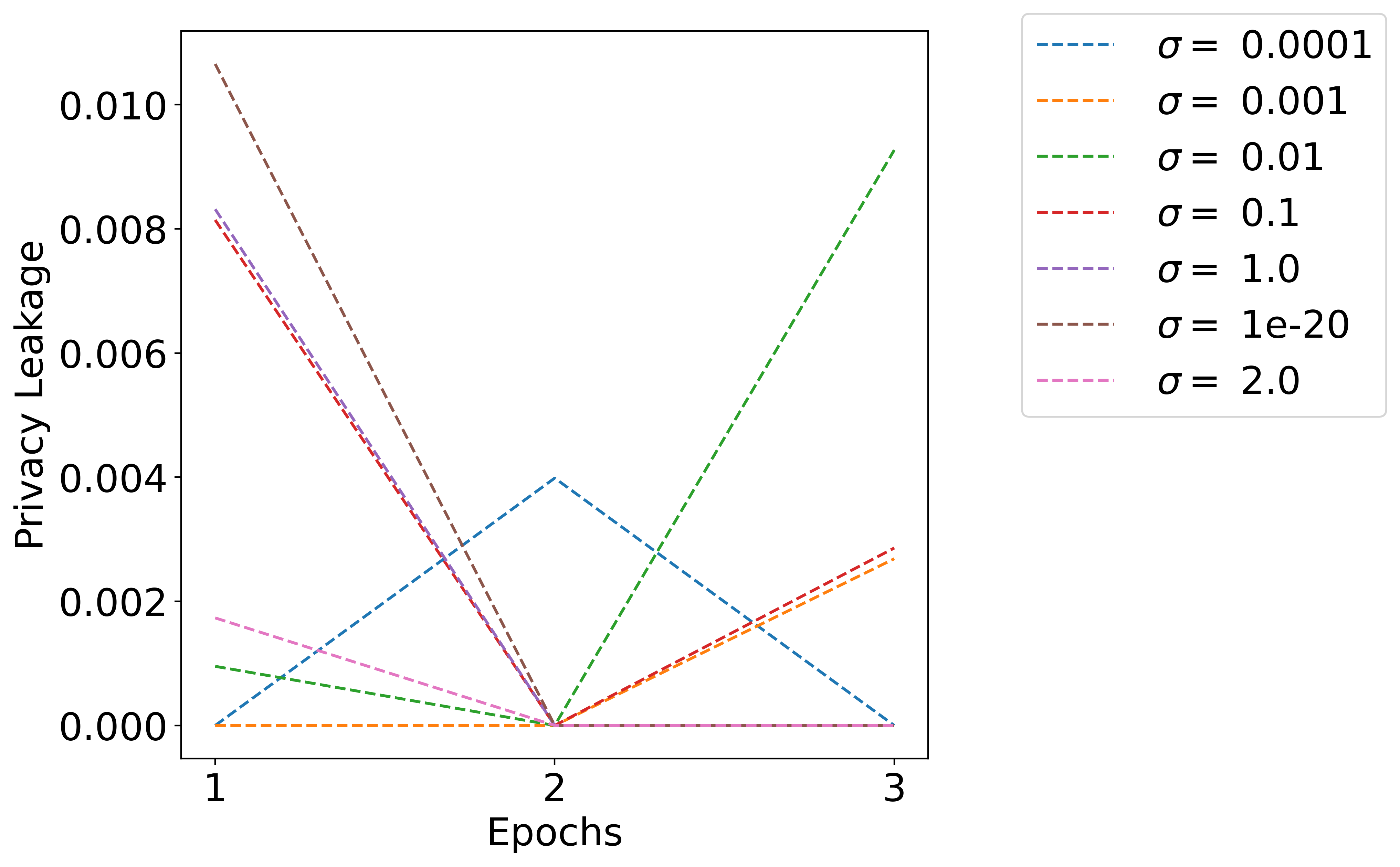}
    \caption{Group privacy leakage for Distil\textsc{bert} models with varying $\sigma$ values using MIMIC-III data. Leakage obtained using patient level black box attack (P-BBA). Gradient clip value is 1, group samples are limited to 50.}
    \label{fig:mimic_bb_patient_distilbert_priv}
\end{figure*}

Figures \ref{fig:mimic_bb_patient_bert_priv}, \ref{fig:mimic_bb_patient_distilbert_priv} and \ref{fig:mimic_bb_patient_gpt2_priv} show patient level black box (P-BBA) attack privacy leakage for \textsc{bert}, Distil\textsc{bert} and \textsc{gpt2} models. 

\begin{figure*}
    \centering
    \includegraphics[width=0.9\columnwidth]{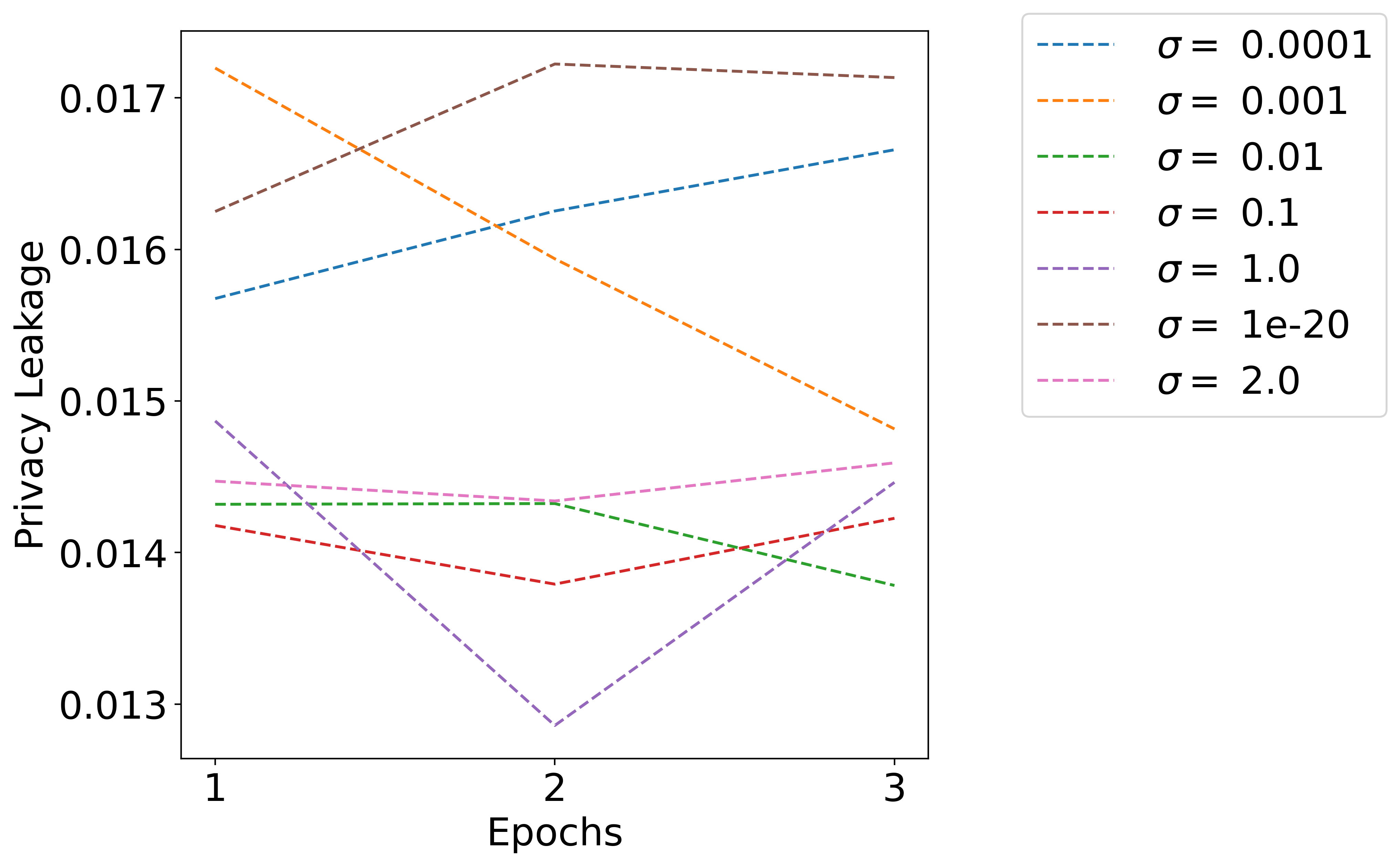}
    \caption{Privacy leakage for \textsc{gpt2} models with varying $\sigma$ values using MIMIC-III data. Leakage obtained using Attention based white box attack (S-AWBA). Gradient clip value is 1, group samples are limited to 50.}
    \label{fig:mimic_wb_attent_sample_gpt2_priv}
\end{figure*}

\begin{figure*}
    \centering
    \includegraphics[width=0.9\columnwidth]{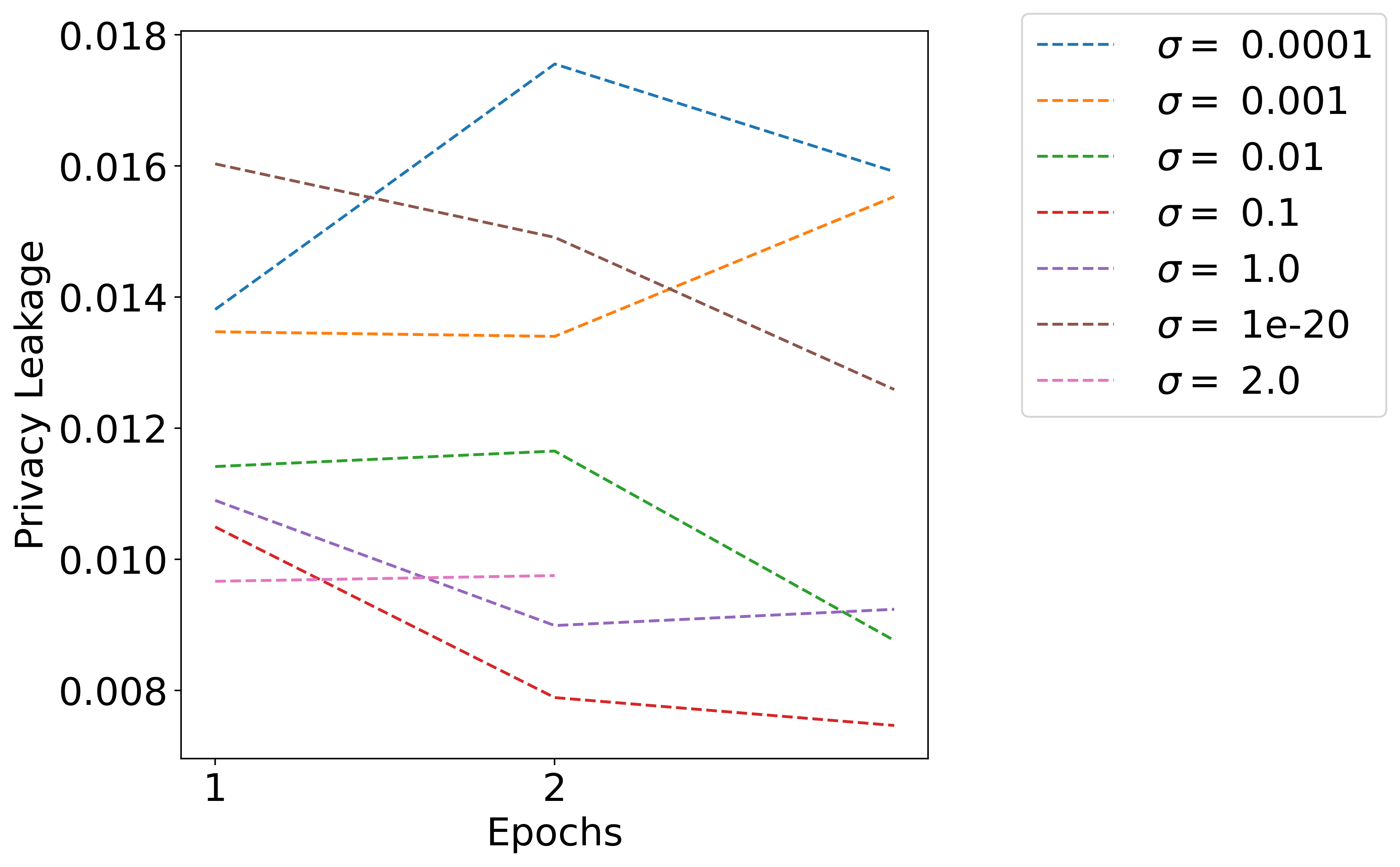}
    \caption{Privacy leakage for \textsc{bert} models with varying $\sigma$ values using MIMIC-III data. Leakage obtained using Attention based white box attack (S-AWBA). Gradient clip value is 1, group samples are limited to 50.}
    \label{fig:mimic_wb_attent_sample_bert_priv}
\end{figure*}

\begin{figure*}
    \centering
    \includegraphics[width=0.9\columnwidth]{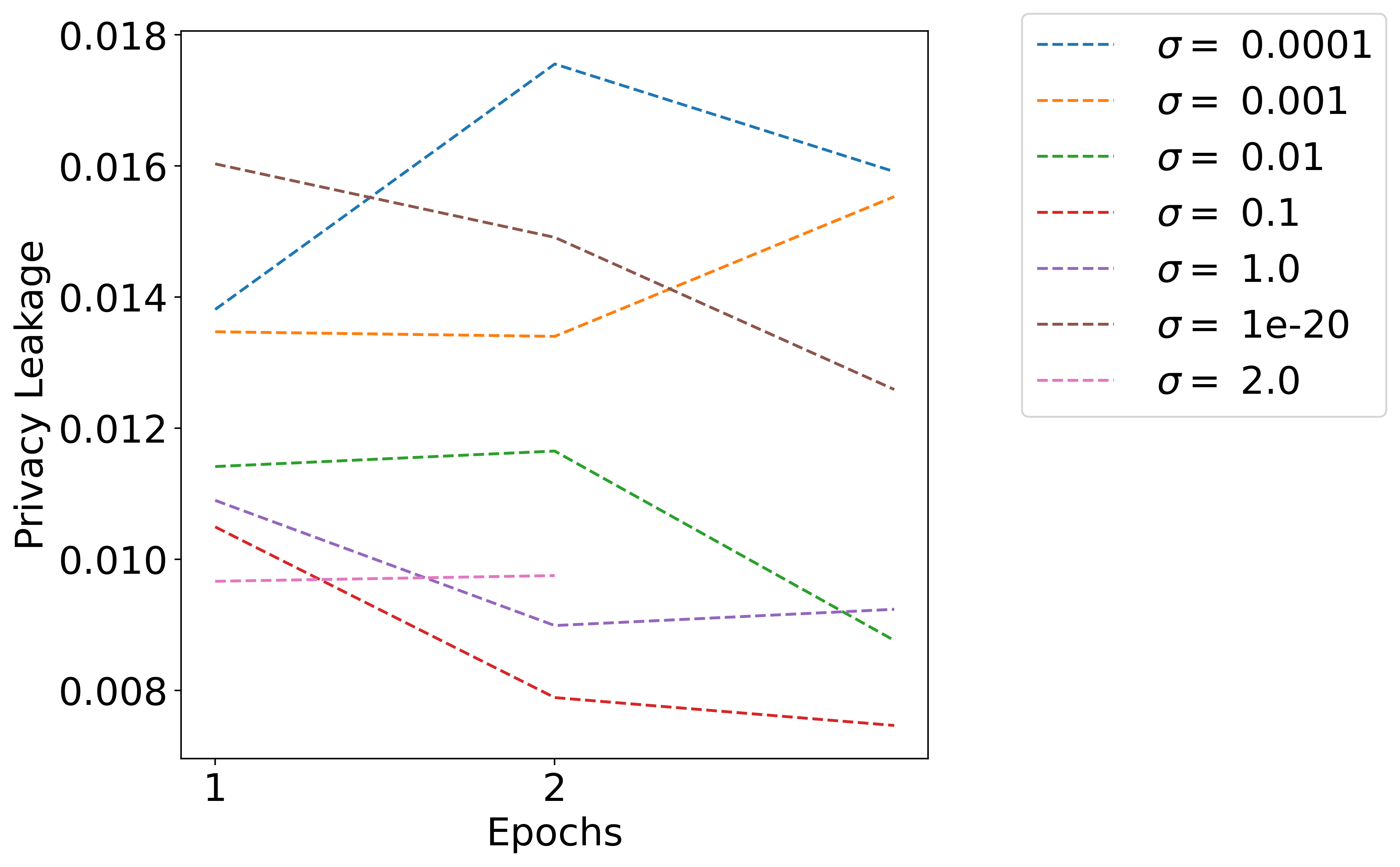}
    \caption{Privacy leakage for Distil\textsc{bert} models with varying $\sigma$ values using MIMIC-III data. Leakage obtained using Attention based white box attack (S-AWBA). Gradient clip value is 1, group samples are limited to 50.}
    \label{fig:mimic_wb_attent_sample_distilbert_priv}
\end{figure*}

Figures \ref{fig:mimic_wb_attent_sample_bert_priv}, \ref{fig:mimic_wb_attent_sample_distilbert_priv} and \ref{fig:mimic_wb_attent_sample_gpt2_priv} show S-AWBA attack privacy leakage for \textsc{bert}, Distil\textsc{bert} and \textsc{gpt2} models.

\begin{figure*}
    \centering
    \includegraphics[width=0.9\columnwidth]{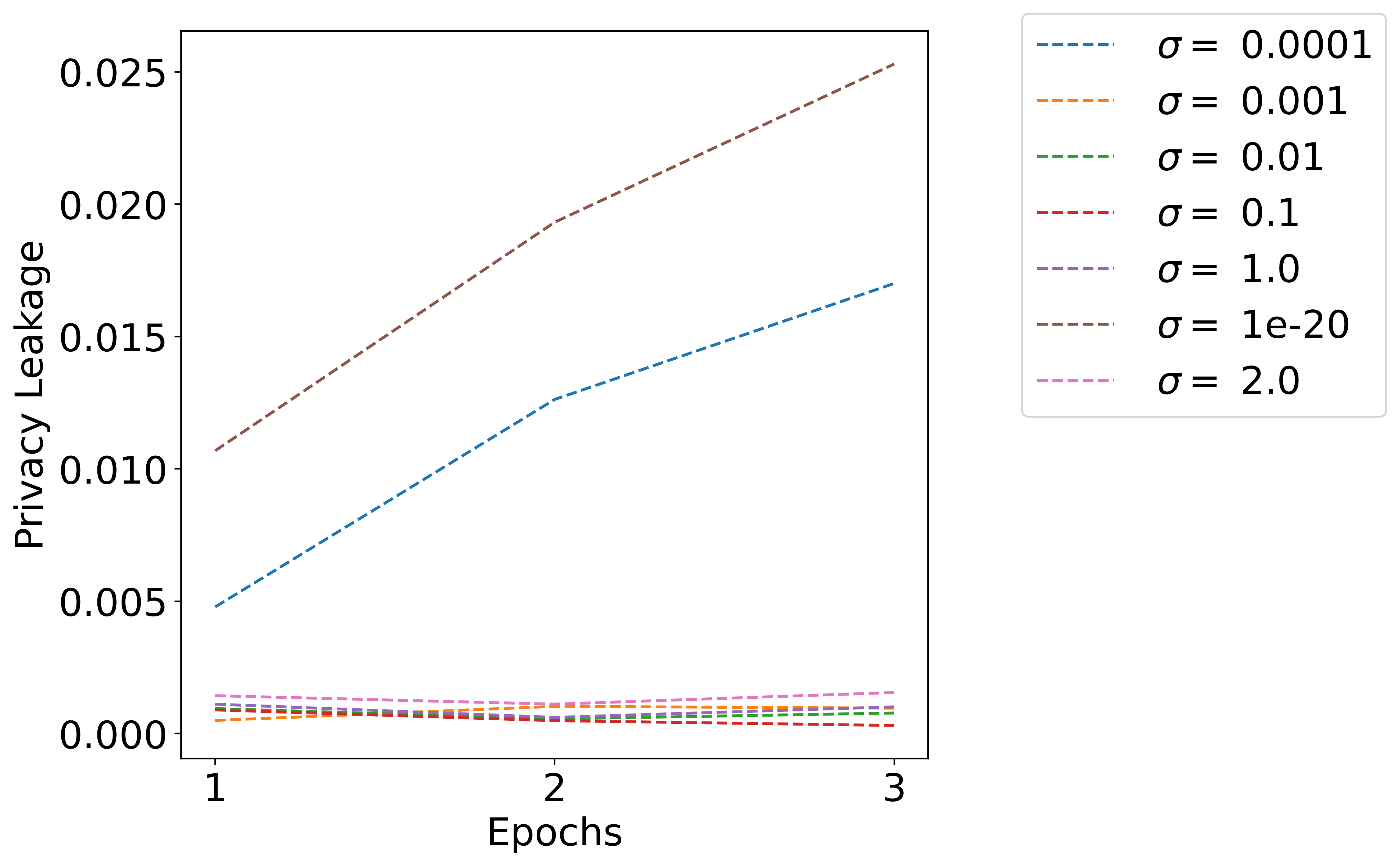}
    \caption{Privacy leakage for \textsc{gpt2} models with varying $\sigma$ values using MIMIC-III data. Leakage obtained using Gradient based white box attack (S-GWBA). Gradient clip value is 1, group samples are limited to 50.}
    \label{fig:mimic_wb_grad_sample_gpt2_priv}
\end{figure*}

\begin{figure*}
    \centering
    \includegraphics[width=0.9\columnwidth]{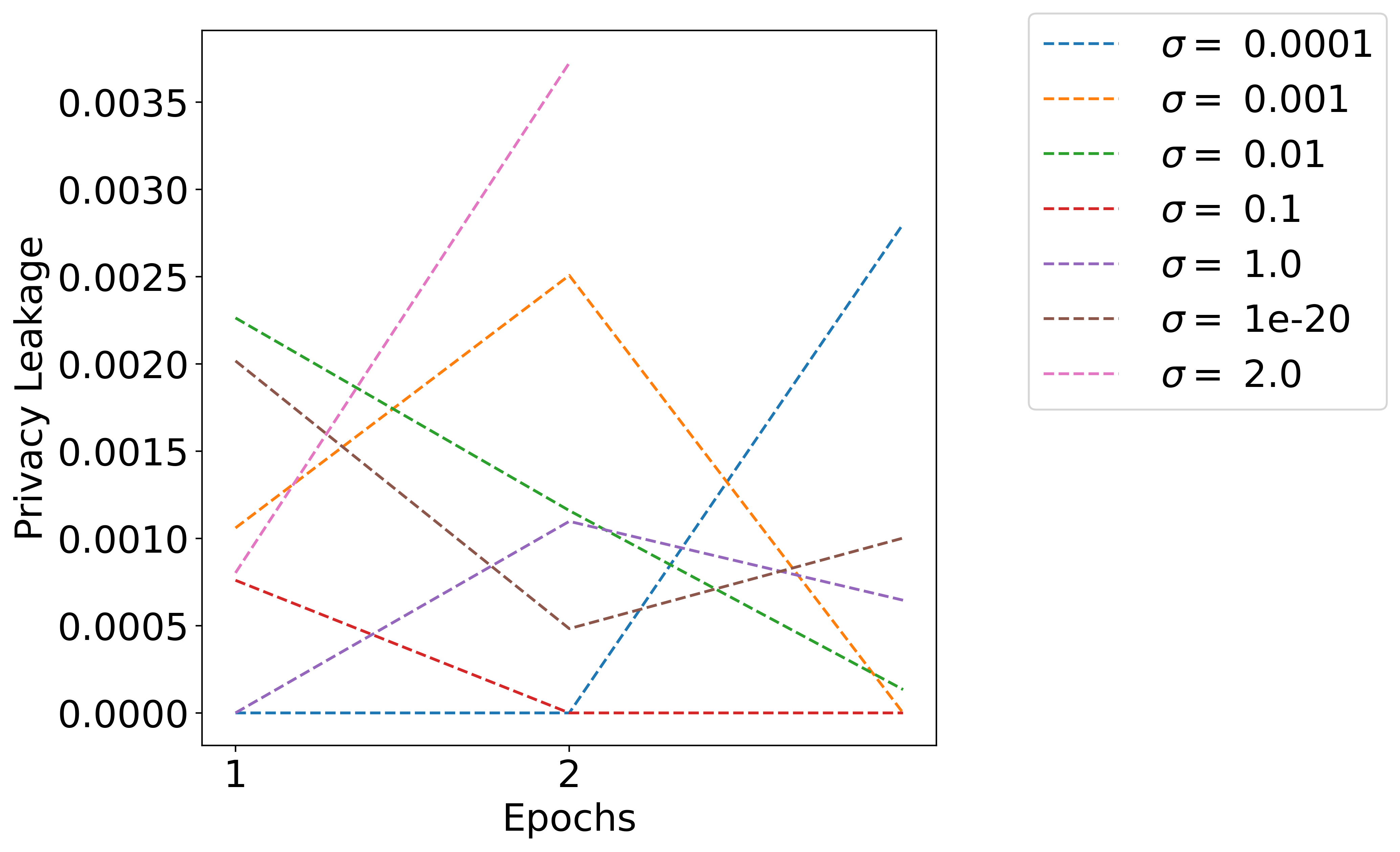}
    \caption{Privacy leakage for \textsc{bert} models with varying $\sigma$ values using MIMIC-III data. Leakage obtained using Gradient based white box attack (S-GWBA). Gradient clip value is 1, group samples are limited to 50.}
    \label{fig:mimic_wb_grad_sample_bert_priv}
\end{figure*}

\begin{figure*}
    \centering
    \includegraphics[width=0.9\columnwidth]{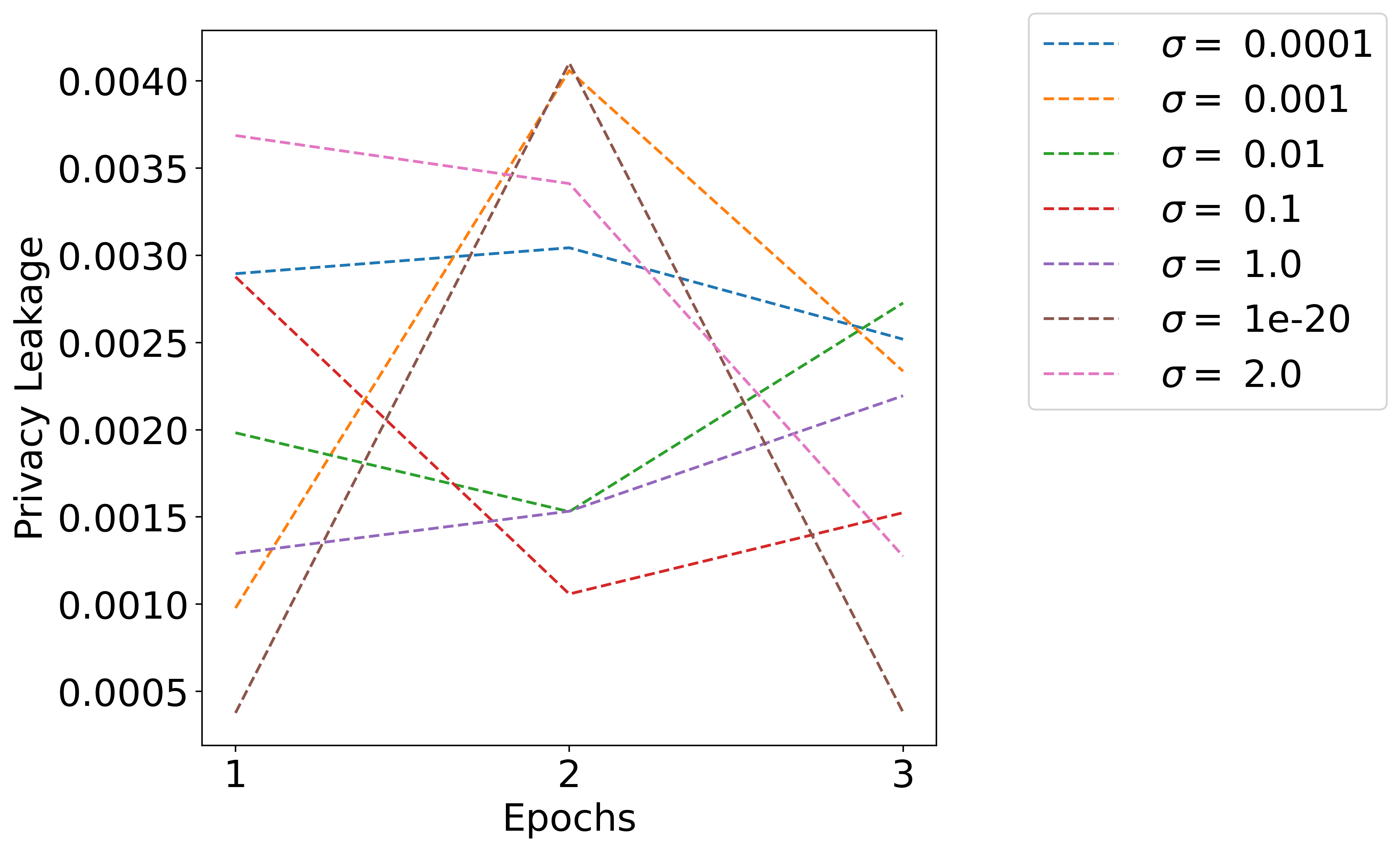}
    \caption{Privacy leakage for Distil\textsc{bert} models with varying $\sigma$ values using MIMIC-III data. Leakage obtained using Gradient based white box attack (S-GWBA). Gradient clip value is 1, group samples are limited to 50.}
    \label{fig:mimic_wb_grad_sample_distilbert_priv}
\end{figure*}

Figures \ref{fig:mimic_wb_grad_sample_bert_priv}, \ref{fig:mimic_wb_grad_sample_distilbert_priv} and \ref{fig:mimic_wb_grad_sample_gpt2_priv} show S-GWBA attack privacy leakage for \textsc{bert}, Distil\textsc{bert} and \textsc{gpt2} models.

\subsubsection{LM Loss Results}
\begin{figure*}
    \centering
    \includegraphics[width=0.9\columnwidth]{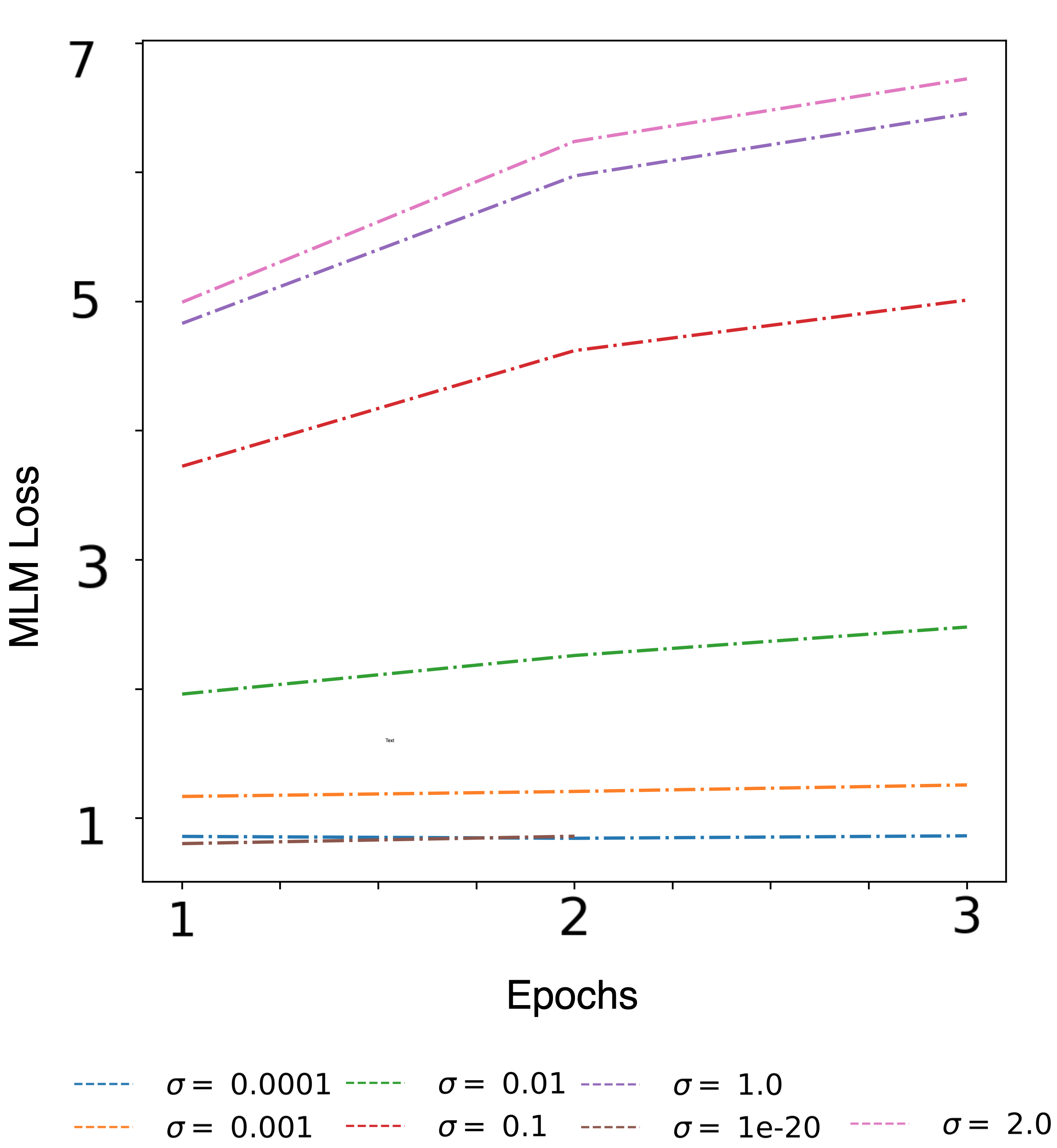}
    \caption{Test data LM loss for DP \textsc{bert} models with varying $\sigma$ values using MIMIC-III data. Gradient clip value is 1, group samples are limited to 50. Loss of \textsc{bert}-base-cased model that is not trained on clinical data is 3.49. All DP models with less than 3.49 loss plots have increased model utility due to DP training. Non-DP CLM model loss is 0.60,0.57,0.55,0.53 for epoch 1 - 4. Non-DP mlm loss is lower than all DP models.}
    \label{fig:mimic_perplexity_bert_priv}
\end{figure*}

\begin{figure*}
    \centering
    \includegraphics[width=0.9\columnwidth]{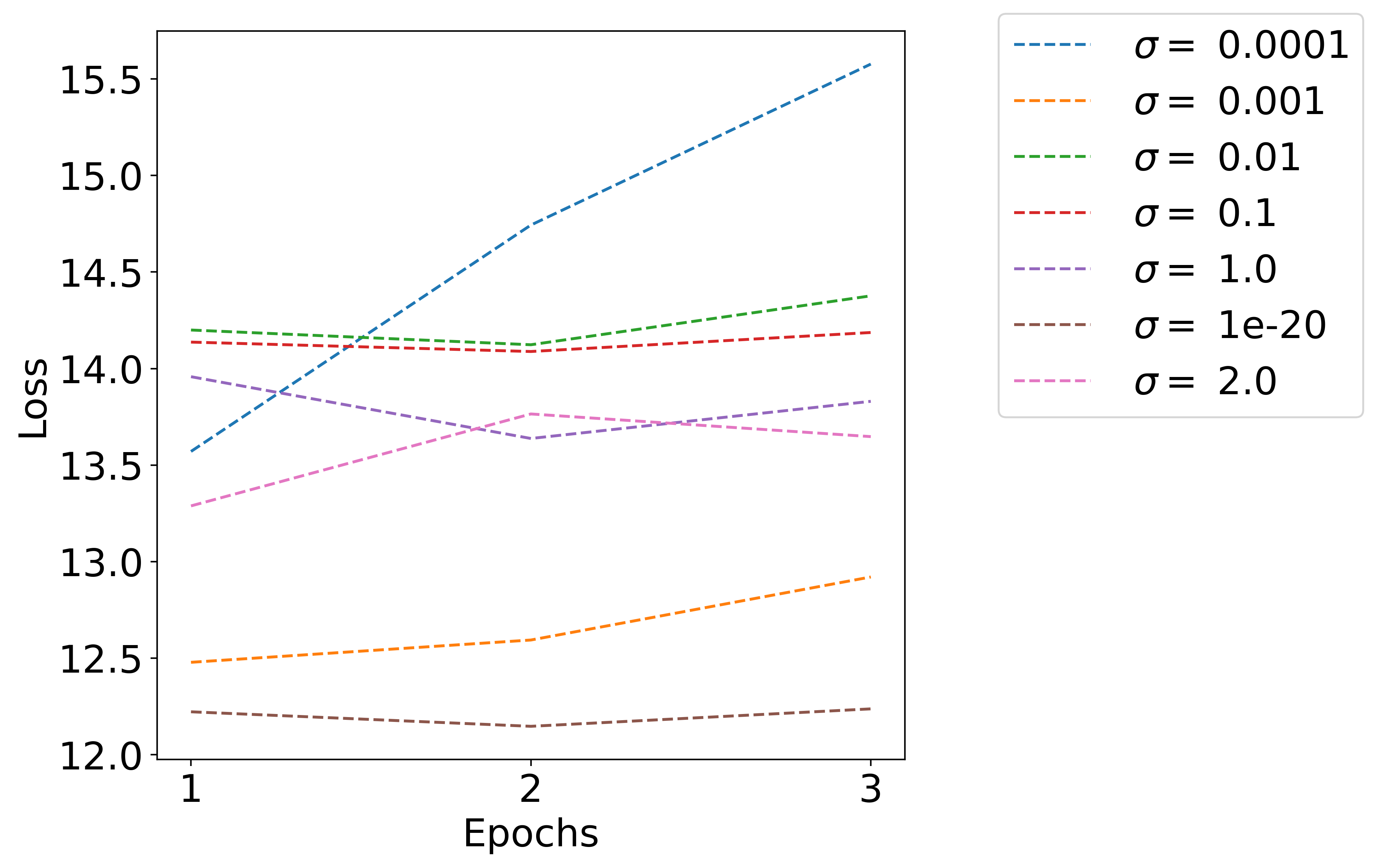}
    \caption{Test data LM loss for DP Distil\textsc{bert} models with varying $\sigma$ values using MIMIC-III data. Gradient clip value is 1, group samples are limited to 50. Loss of Distil\textsc{bert}-base model that is not trained on clinical data is 3.49. All DP models with less than 3.49 loss plots have increased model utility due to DP training. Non-DP CLM model loss is 0.63,0.59,0.58,0.58 for epoch 1 - 4. Non-DP mlm loss is lower than all DP models.}
    \label{fig:mimic_perplexity_distilbert_priv}
\end{figure*}

\begin{figure*}
    \centering
    \includegraphics[width=0.9\columnwidth]{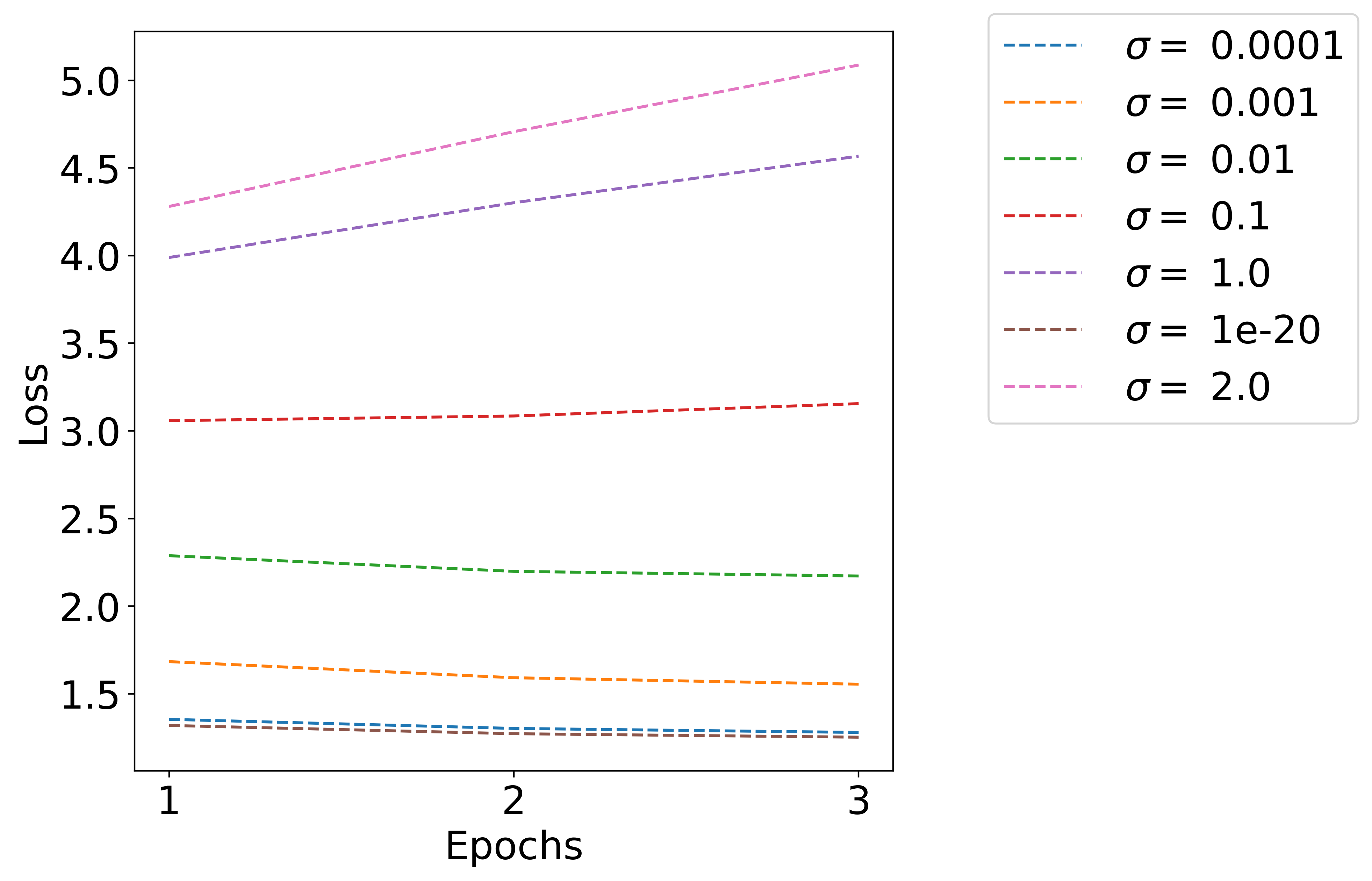}
    \caption{Test data LM loss for \textsc{gpt2} models with varying $\sigma$ values using MIMIC-III data. Gradient clip value is 1, group samples are limited to 50. Loss of \textsc{gpt2}-base model that is not trained on clinical data is 3.49. All DP models with less than 3.49 loss plots have increased model utility due to DP training. Non-DP CLM model loss is 1.2,1.19,1.17,1.16 for epoch 1 - 4.}
    \label{fig:mimic_perplexity_gpt_priv}
\end{figure*}

Figure \ref{fig:mimic_perplexity_bert_priv} and \ref{fig:mimic_perplexity_distilbert_priv} provide masked language modeling loss for all\textsc{bert} and distil\textsc{bert}DP models, trained on MIMIC-III. Figure \ref{fig:mimic_perplexity_gpt_priv} shows autoregressive LM loss for \textsc{gpt2} DP models. 

\subsubsection{Privacy Leakage based on Patient Profiles}
\begin{figure*}
    \centering
    \subfigure[]{\includegraphics[width=0.4\textwidth]{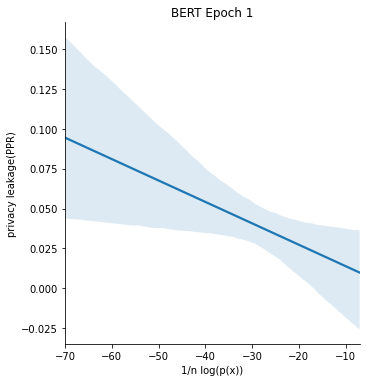}}
    \subfigure[]{\includegraphics[width=0.4\textwidth]{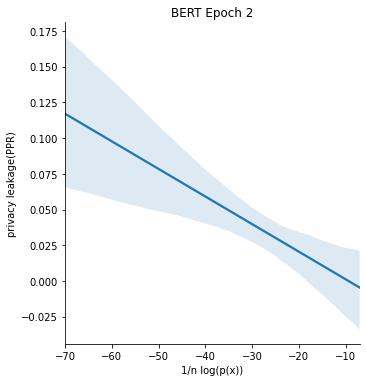}}
    \subfigure[]{\includegraphics[width=0.4\textwidth]{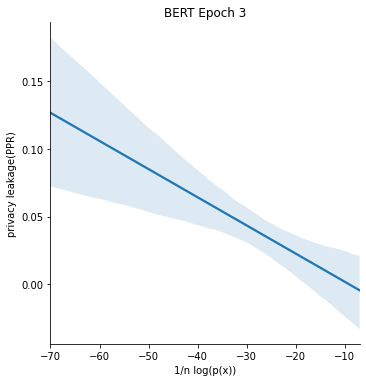}}
    \subfigure[]{\includegraphics[width=0.4\textwidth]{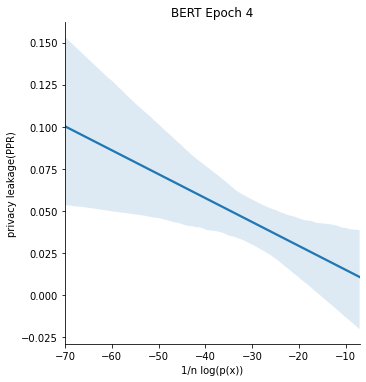}}
    \caption{Privacy leakage vs log-normalized probability plots for Non-DP trained \textsc{bert} models on MIMIC-III data. We see the negative correlation between probability of disease profile (admission-level) and privacy leakage.}
    \label{fig:mimic_nondp_bert_epochs}
\end{figure*}

\begin{figure*}
    \centering
    \subfigure[]{\includegraphics[width=0.4\textwidth]{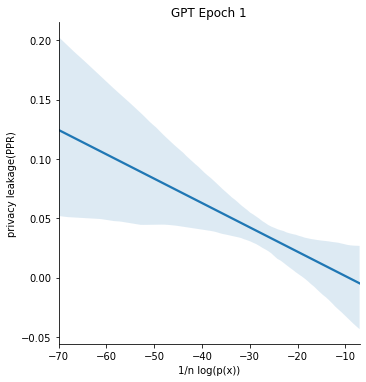}}
    \subfigure[]{\includegraphics[width=0.4\textwidth]{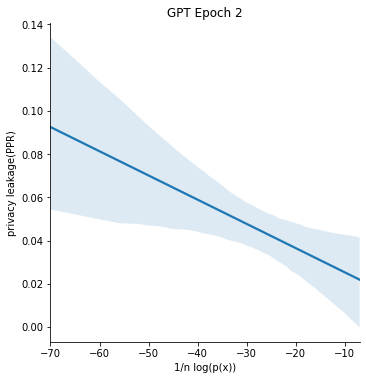}}
    \subfigure[]{\includegraphics[width=0.4\textwidth]{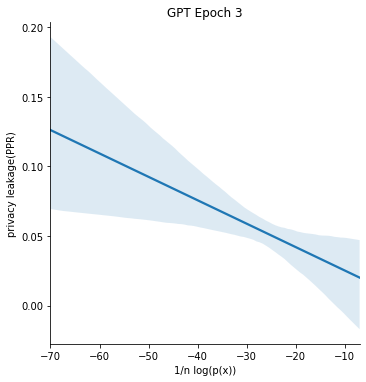}}
    \subfigure[]{\includegraphics[width=0.4\textwidth]{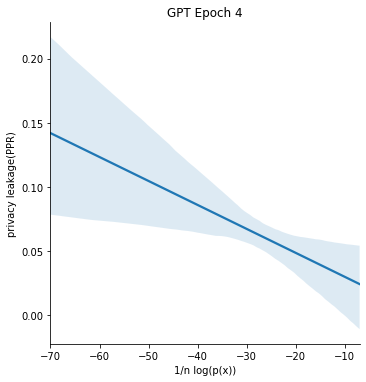}}
    \caption{Privacy leakage vs log-normalized probability plots for Non-DP trained \textsc{gpt2} models on MIMIC-III data. We see the negative correlation between probability of disease profile (admission-level) and privacy leakage.}
    \label{fig:mimic_nondp_gpt_epochs}
\end{figure*}

\begin{figure*}
    \centering
    \subfigure[]{\includegraphics[width=0.4\textwidth]{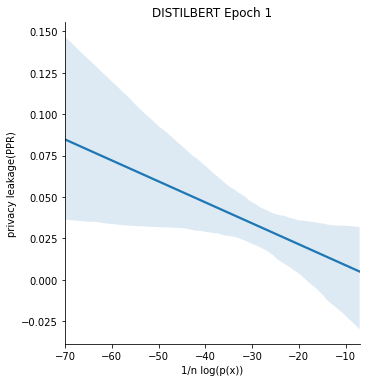}}
    \subfigure[]{\includegraphics[width=0.4\textwidth]{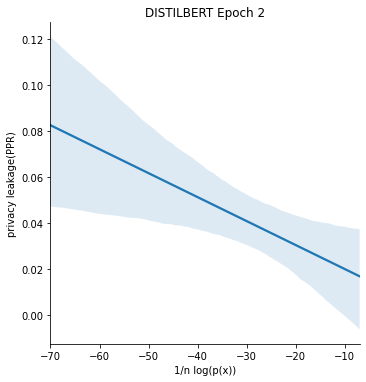}}
    \subfigure[]{\includegraphics[width=0.4\textwidth]{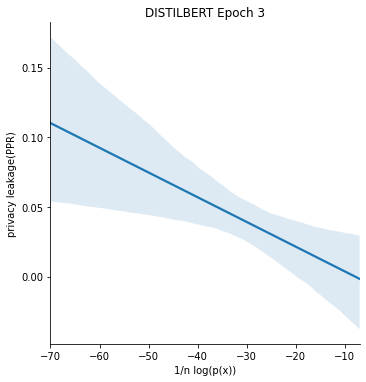}}
    \subfigure[]{\includegraphics[width=0.4\textwidth]{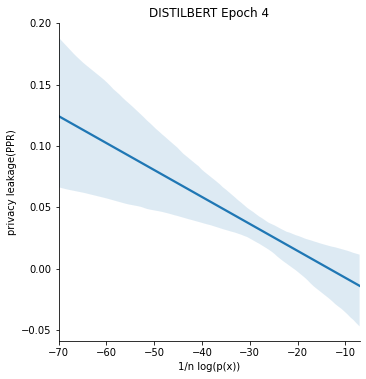}}
    \caption{Privacy leakage vs log-normalized probability plots for Non-DP trained Distil\textsc{bert} models on MIMIC-III data. We see the negative correlation between probability of disease profile (admission-level) and privacy leakage.}
    \label{fig:mimic_nondp_distilbert_epochs}
\end{figure*}

In this section we provide detailed results for correlations between patient privacy profiles and privacy leakage. We divide MIMIC-III admissions and patients into buckets based on the log normalized probability of their disease profile. The probability of an admission or patient's disease profile is obtain using the procedure outlined in Section \ref{app:disease_profile_calculation}. Plots of correlation between a bucket's (admission-level) privacy leakage and its patient disease probability are in Figure \ref{fig:mimic_nondp_bert_epochs}, \ref{fig:mimic_nondp_gpt_epochs}  and \ref{fig:mimic_nondp_distilbert_epochs}. Trends for patient level disease profile and privacy leakage are same as shown in the aforementioned plots.

\subsubsection{MedNLI Results for MIMIC-III Models}
In Table~\ref{tab:mnli_bert_accuracy}, we see that Non-DP \textsc{bert}model achieves an accuracy of $0.8023$ on MedNLI dataset \cite{romanov2018lessons} and BERT-base-cased achieved a performance of $0.776$ per \cite{alsentzer2019publicly}. All the DP trained model achieved a performance higher than BERT-base-cased model and quite close to Non-DP trained \textsc{bert}model which shows the utility of fine-tuning models with differentially private training technique.

\subsection{Detailed Results for UMM Dataset}
% \bhanu{todo}

\begin{figure*}
    \centering
    \includegraphics[width=1\textwidth]{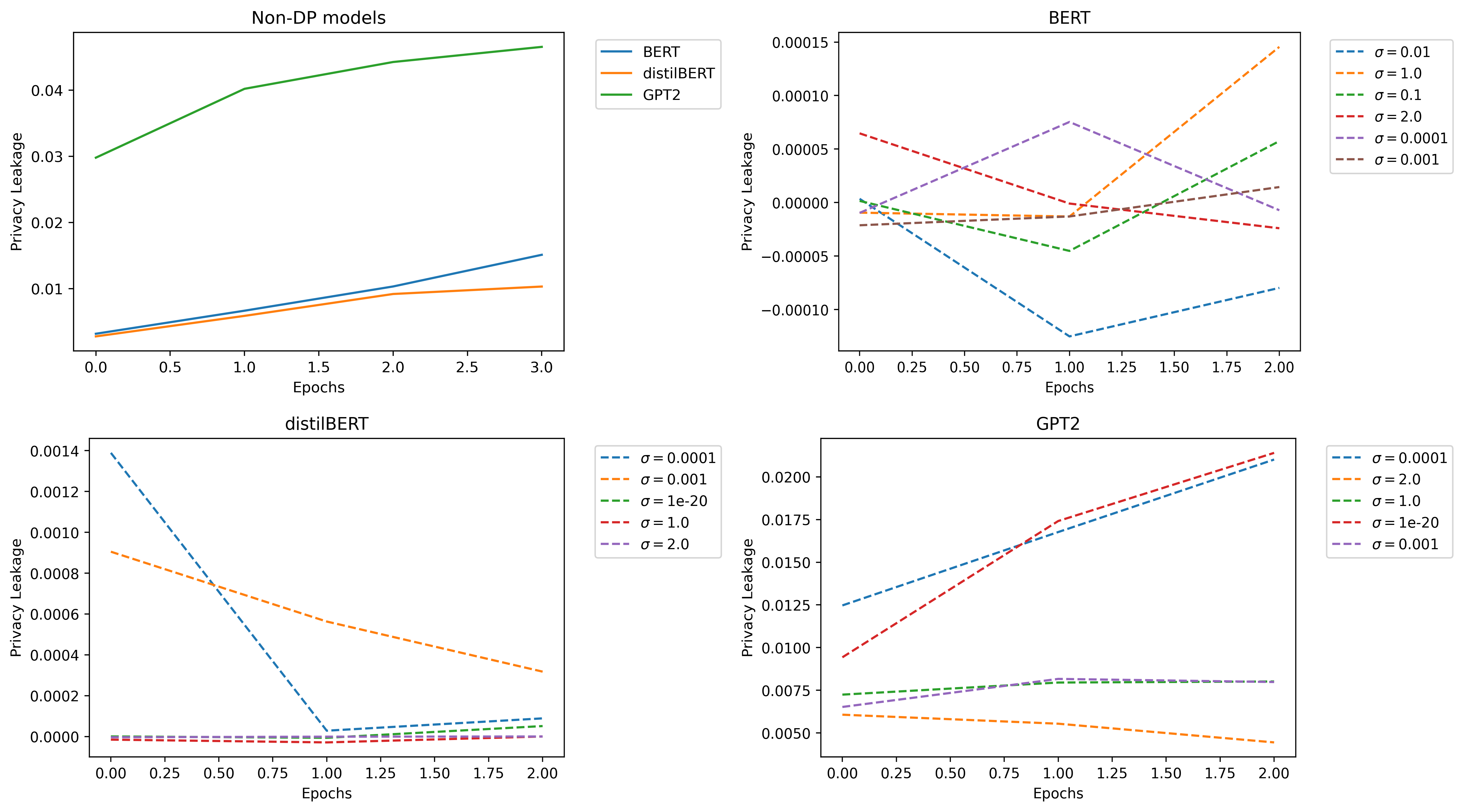}
    \caption{Privacy leakage using Gradient-based white box attack (S-GWBA) for Hospital 2 at sample-level. Models shown are Non-DP and DP trained variants of \textsc{bert}, Distil\textsc{bert}and \textsc{gpt2}. For all DP models, the Gradient clip value is set to 1 and group samples are limited to 50.}
    \label{fig:wb_lr}
\end{figure*}

We see similar privacy leakage pattern for UMM Dataset as MIMIC Dataset. We observe highest privacy leakage for Non-DP \textsc{gpt2} model using gradient-based white box attack (S-GWBA) as shown in Fig.~\ref{fig:wb_lr}.

\begin{figure*}
    \centering
    \includegraphics[width=1\textwidth]{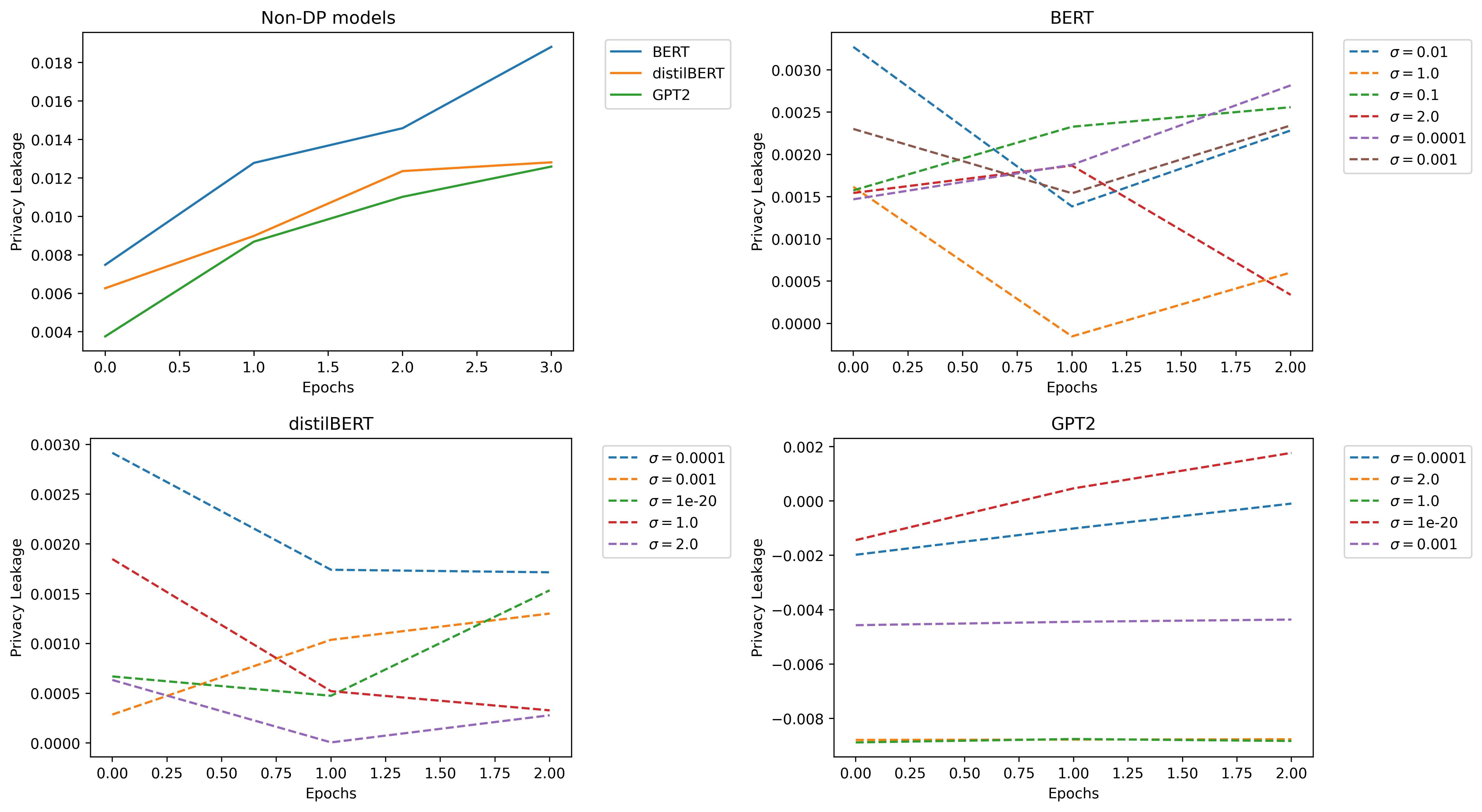}
    \caption{Sample-level Black box attack (S-BBA) privacy leakage using Hospital 2 data  for Non-DP and DP trained variants of \textsc{bert}, Distil\textsc{bert}and \textsc{gpt2}. For all DP models, the Gradient clip value is set to 1 and group samples are limited to 50.}
    \label{fig:bb_sample}
\end{figure*}

Fig.~\ref{fig:bb_sample} show sample level black box attack (S-BBA) privacy leakages for \textsc{bert}, Distil\textsc{bert} and \textsc{gpt2} models. 

\begin{figure*}
    \centering
    \includegraphics[width=1\textwidth]{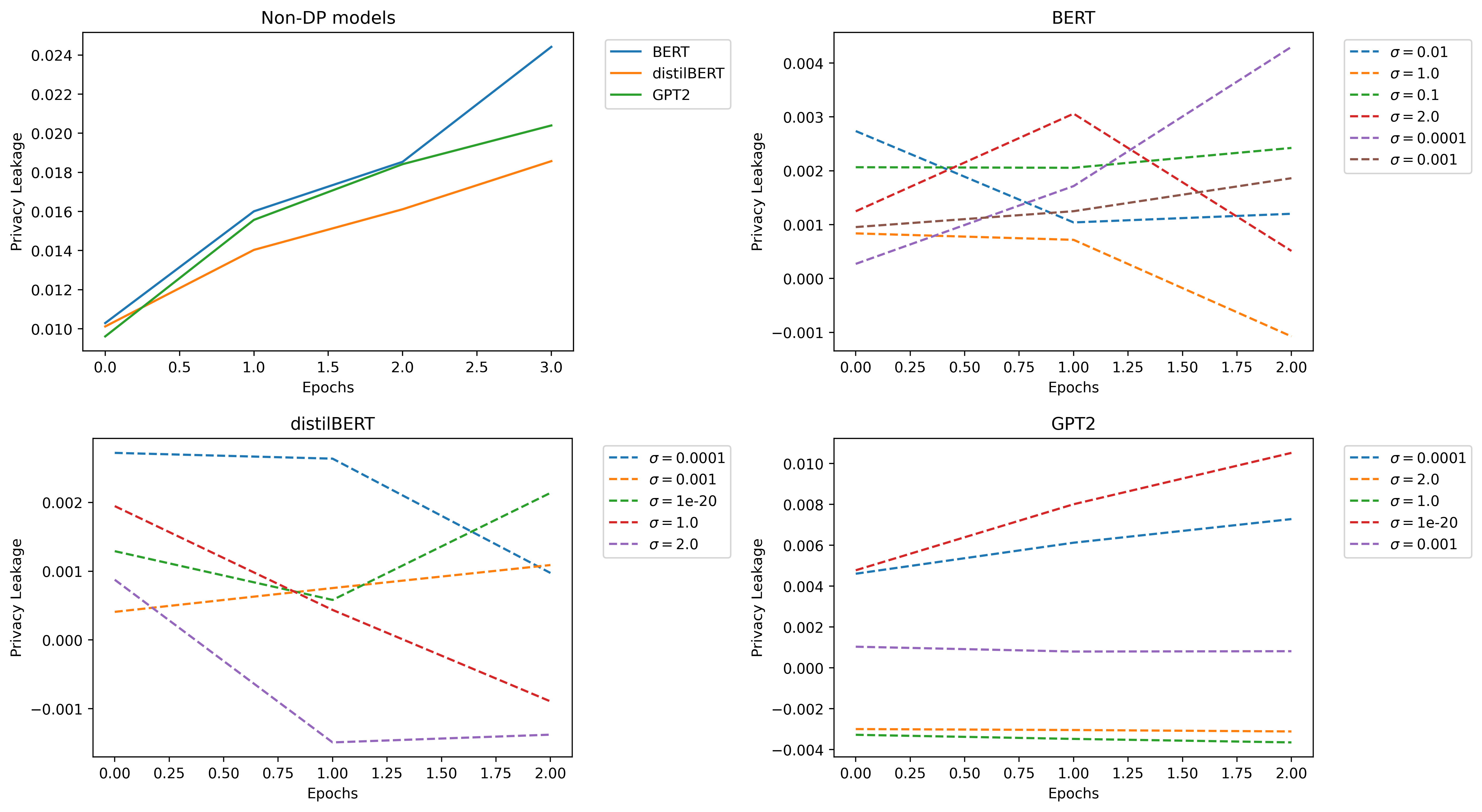}
    \caption{Admission-level Black box attack (A-BBA) group privacy leakage using Hospital 2 data for Non-DP and DP trained variants of \textsc{bert}, Distil\textsc{bert}and \textsc{gpt2}. For all DP models, the Gradient clip value is set to 1 and group samples are limited to 50.}
    \label{fig:bb_hadm}
\end{figure*}

Fig.~\ref{fig:bb_hadm} show hospital admission level black box attack (A-BBA) privacy leakages for \textsc{bert}, Distil\textsc{bert} and \textsc{gpt2} models. 

\begin{figure*}
    \centering
    \includegraphics[width=1\textwidth]{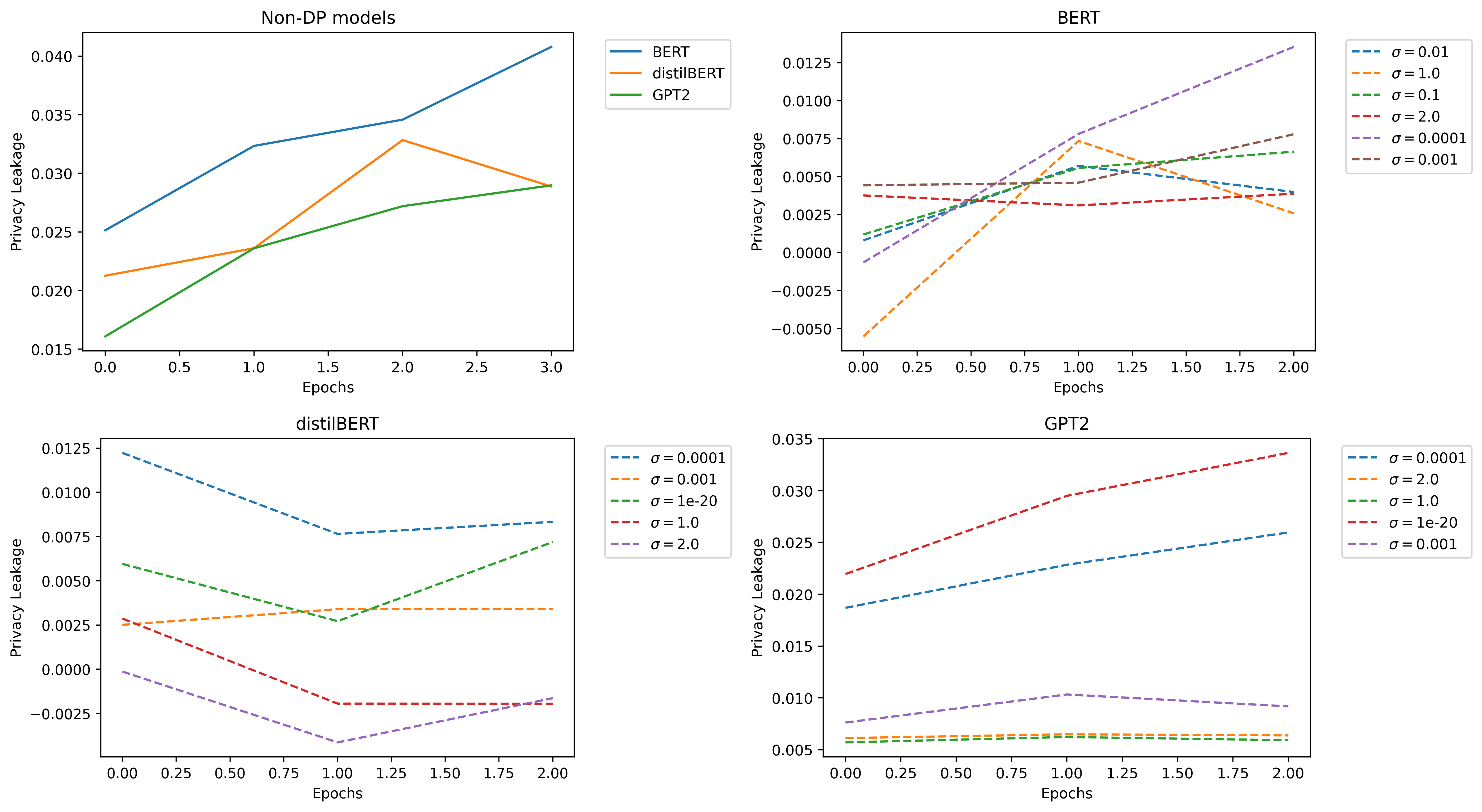}
    \caption{Patient-level Black box attack (P-BBA) group privacy leakage using Hospital 2 data  for Non-DP and DP trained variants of \textsc{bert}, Distil\textsc{bert}and \textsc{gpt2}. For all DP models, the Gradient clip value is set to 1 and group samples are limited to 50.}
    \label{fig:bb_pat}
\end{figure*}

Fig.~\ref{fig:bb_pat} show patient level black box attack (P-BBA) privacy leakages for \textsc{bert}, Distil\textsc{bert} and \textsc{gpt2} models. 

\begin{figure*}
    \centering
    \includegraphics[width=1\textwidth]{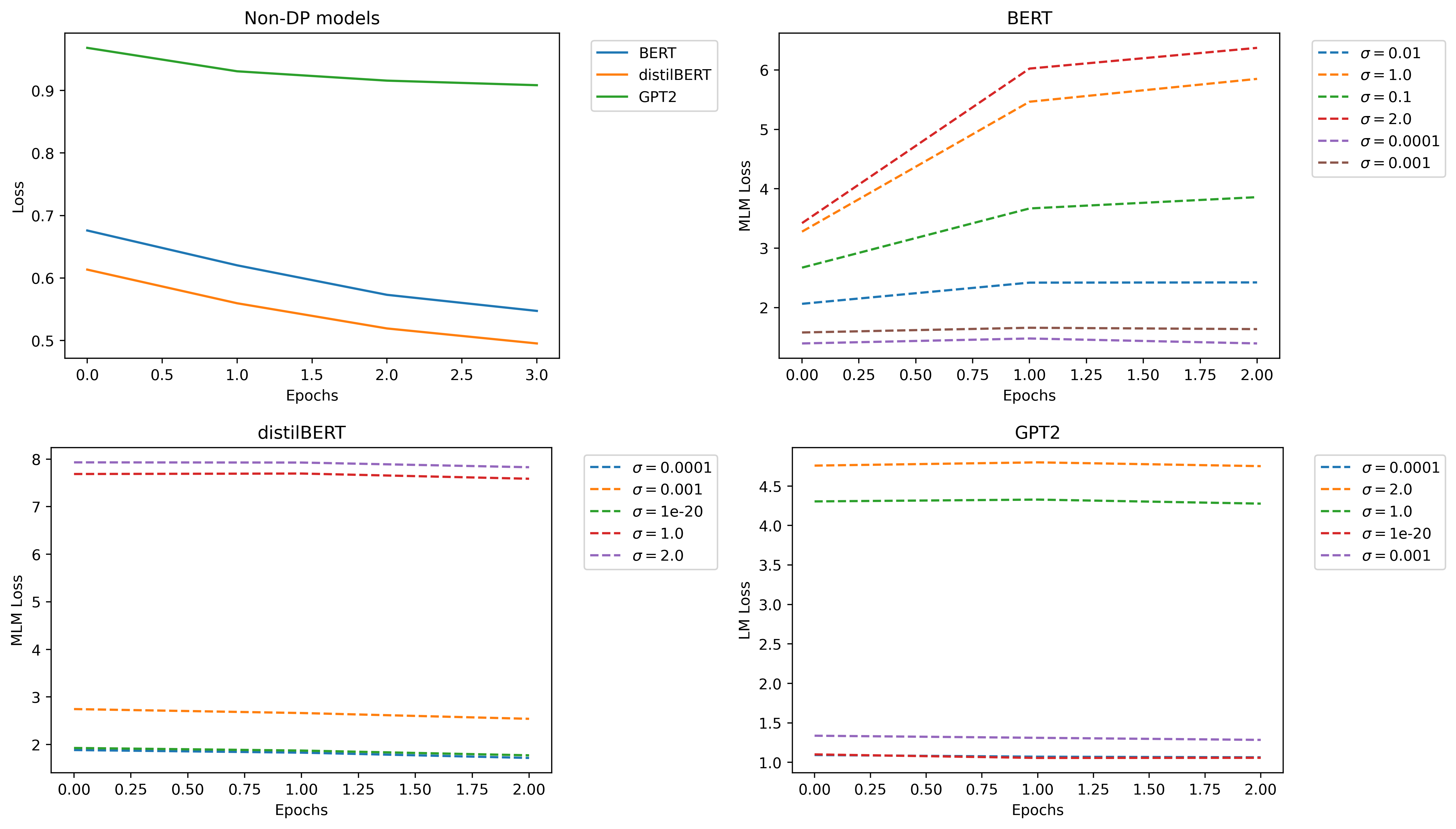}
    \caption{Negative Log Likelihood Loss for Non-DP CLMs and their DP trained variants with varying $\sigma$ values. For all DP models, the Gradient clip value is set to 1 and group samples are limited to 50.}
    \label{fig:perp__}
\end{figure*}

Fig.~\ref{fig:perp__} show negative log likelihood loss for \textsc{bert}, Distil\textsc{bert} and \textsc{gpt2} models.

\begin{figure*}
    \centering
    \includegraphics[width=1\textwidth]{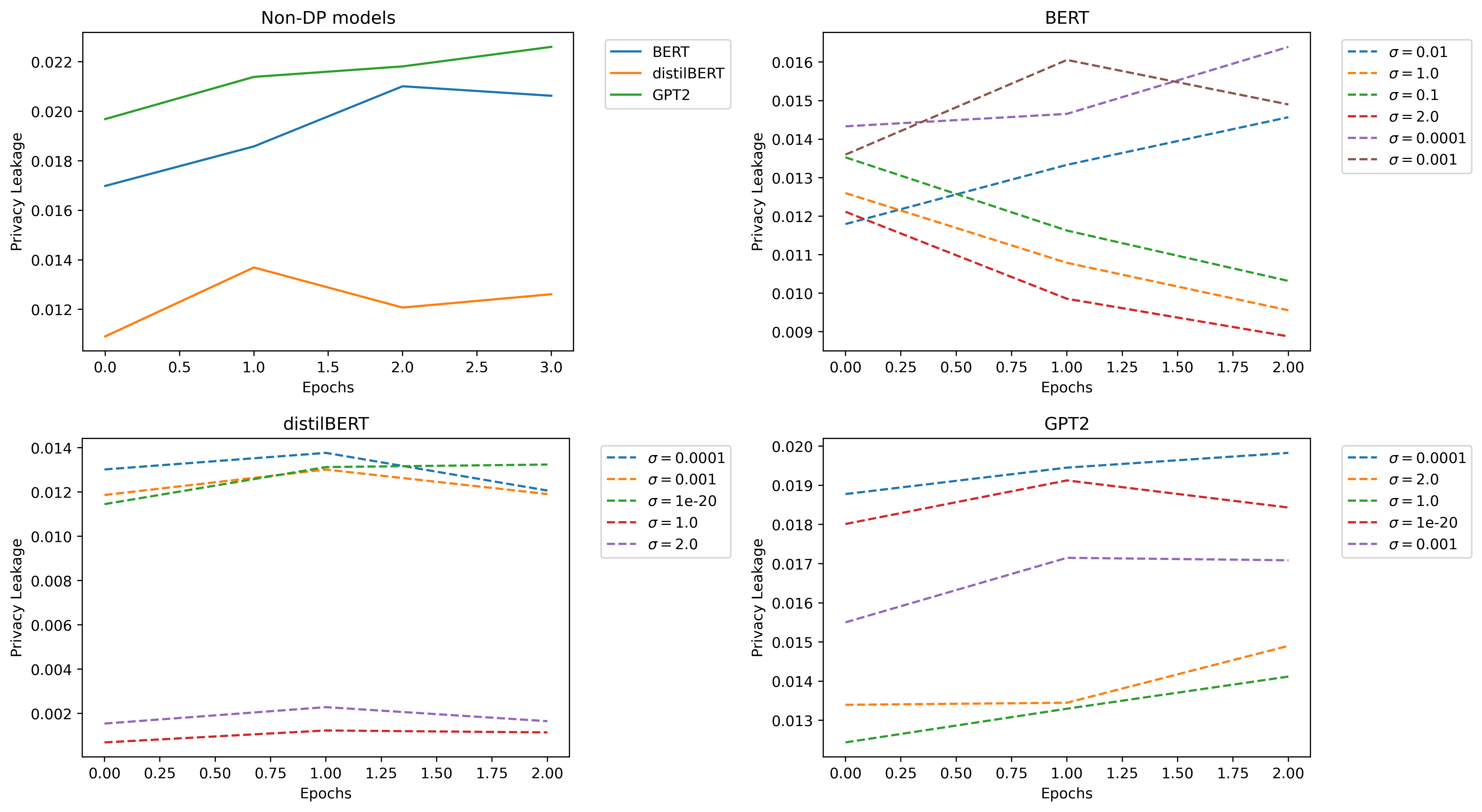}
    \caption{Privacy leakage using Attention-based white-box attack (S-AWBA) for Hospital 2 at sample level. Models shown are Non-DP and DP trained variants of \textsc{bert}, Distil\textsc{bert}and \textsc{gpt2}. For all DP models, the Gradient clip value is set to 1 and group samples are limited to 50.}
    \label{fig:bb_awa}
\end{figure*}

Fig.~\ref{fig:bb_awa} show sample level attention-based white box attack (S-AWBA) privacy leakages for \textsc{bert}, Distil\textsc{bert} and \textsc{gpt2} models. 

Fig.~\ref{fig:wb_lr} show sample level gradient-based white box attack (S-GWBA) privacy leakages for \textsc{bert}, Distil\textsc{bert} and \textsc{gpt2} models. 

% Fig~\ref{fig:bb_ama} and Fig~\ref{fig:wb_grad}

\subsection{Results for VHA  Dataset}
Due to computational and access constraints for VHA  dataset, we were only able to evaluate black-box-attack on \textsc{bert} model for VHA  data. The patient and sample-level black-box privacy leakages are provided in Table \ref{tab:H3}. We see patterns consistent with MIMIC-III and UMM results. Group-level privacy leakage is higher than sample-level leakage, and increasing epochs increases the privacy leakage. 

\subsection{Useful Notations}
\label{appendix:notations}
\emph{LM:} Language Model
\noindent \newline
\emph{CLM:} Clinical Language Model
\noindent \newline
\emph{DP:} Differentially Private
\noindent \newline
\emph{RDP:} R\'enyi Differential Privacy
\noindent \newline
\emph{PL:} Privacy Leakage
\noindent \newline
\emph{OOT}: out-of-training
\noindent \newline
\emph{A-BBA:} Admission level Black Box Attack
\noindent \newline
\emph{S-BBA:} Sample level Black Box Attack
\noindent \newline
\emph{P-BBA:} Patient level Black Box Attack
\noindent \newline
\emph{S-GWBA:} Sample level Gradient-based White Box Attack
\noindent \newline
\emph{S-AWBA:} Sample level Attention-based White Box Attack
\noindent \newline
\emph{MedNLI:} A Natural Language Inference Dataset for the Clinical Domain \cite{romanov2018lessons}

\begin{table*}[!htbp]
\centering
\begin{tabular}{c|l|ccc}
\toprule
\multicolumn{1}{l}{} &          & \multicolumn{3}{c}{\textsc{bert}Accuracy}  \\ \toprule
\multicolumn{1}{l|}{} & $\sigma$    & Epoch 1   & Epoch 2   & Epoch 3       \\ \midrule
Non-DP               &      -    & 0.7922    & 0.7938    & 0.8023         \\ \midrule
\multirow{7}{*}{DP}  & 1e-4 & 0.8095    & 0.7978    & 0.8086          \\
                     & 1e-3 & 0.7903    & 0.8037    & 0.7952      \\
                     & 1e-2 & 0.7933    & 0.8069    & 0.7926          \\
                     & 1e-1 & 0.8055    & 0.7954    & 0.7861          \\
                     & 1.0       & 0.7926    & 0.8023    & 0.7936         \\
                     & 2.0       & 0.7962    & 0.8015    & 0.7929         \\
                     & 1e-20 & 0.7884    & 0.8098    & 0.7896        \\ \bottomrule
\end{tabular}
\caption{Results for all three epochs of DP and Non-DP \textsc{bert}models. Gradient clip value for DP models is set to 1 and group samples are limited to 50. The Non-DP trained \textsc{bert} model at Epoch 3 is similar to clinical\textsc{bert} model \cite{alsentzer2019publicly} which has the reported performance of \textbf{0.808} on MedNLI dataset. The BERT-base-cased model which is not finetuned on MIMIC-III data achieved an accuracy of \textbf{0.776} per \cite{alsentzer2019publicly}.}
\label{tab:mnli_bert_accuracy}
\end{table*}




\begin{table}
\centering
\begin{tabular}{c|c|c}
\hline
\textsc{bert} epoch & S-BBA 1 & P-BBA\\
\hline
1 & 1.27&  0.92 \\
2 & 1.61 & 1.42\\
3 & 1.89 & 2.09\\
4 & 2.03 & 2.91\\
\hline
\end{tabular}
\caption{ Sample and patient-level black-box attack privacy leakage for \textsc{bert} model using VHA  data.}
\label{tab:H3}
\end{table}

\begin{figure}
    \centering
    \includegraphics[width=0.7\columnwidth]{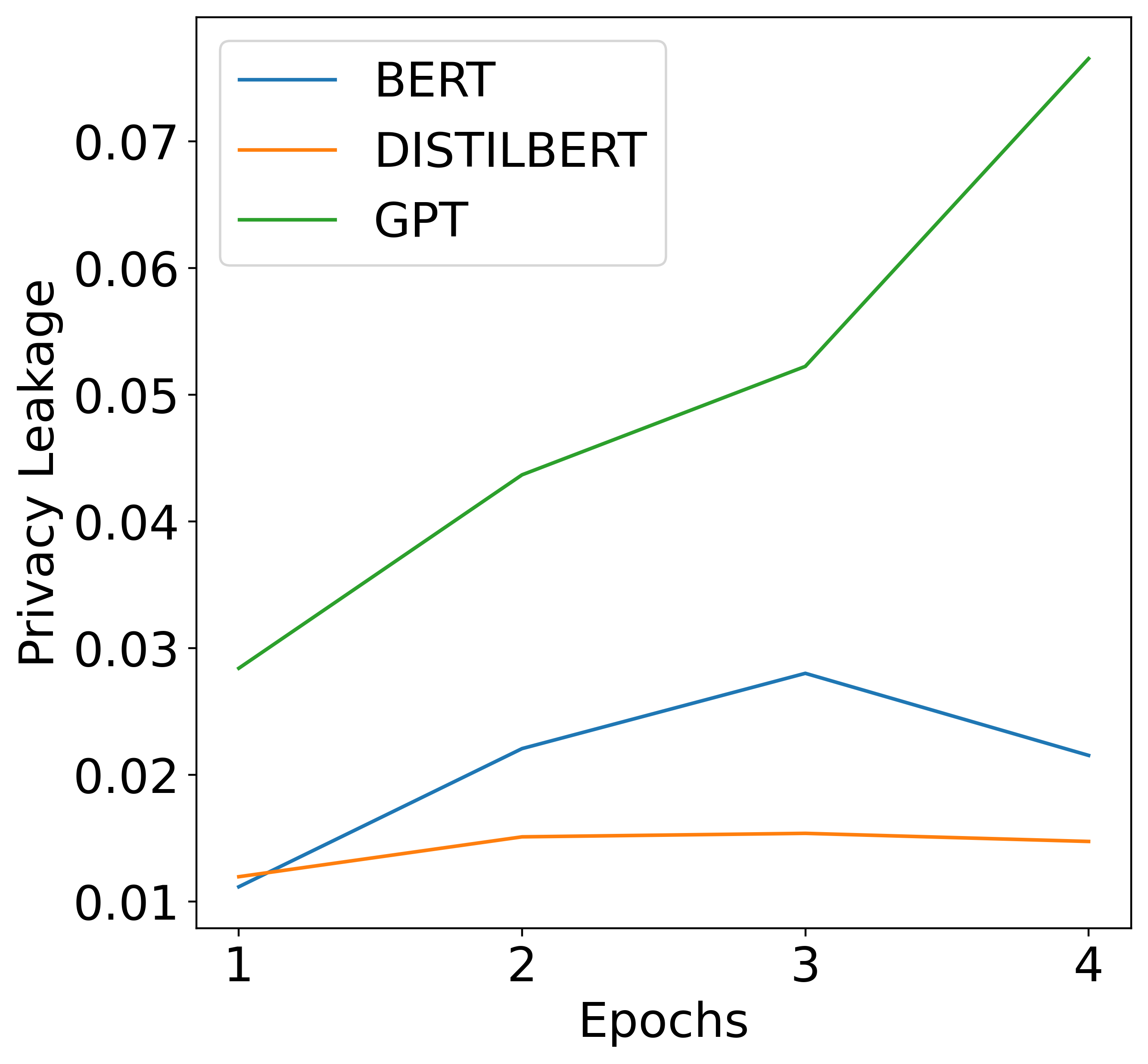}
    \caption{Black box attack privacy leakage for MIMIC-III data with hospital admission level aggregate.}
    \label{fig:mimic_bb_hadm_non_priv}
\end{figure}

\begin{figure}
    \centering
    \includegraphics[width=0.7\columnwidth]{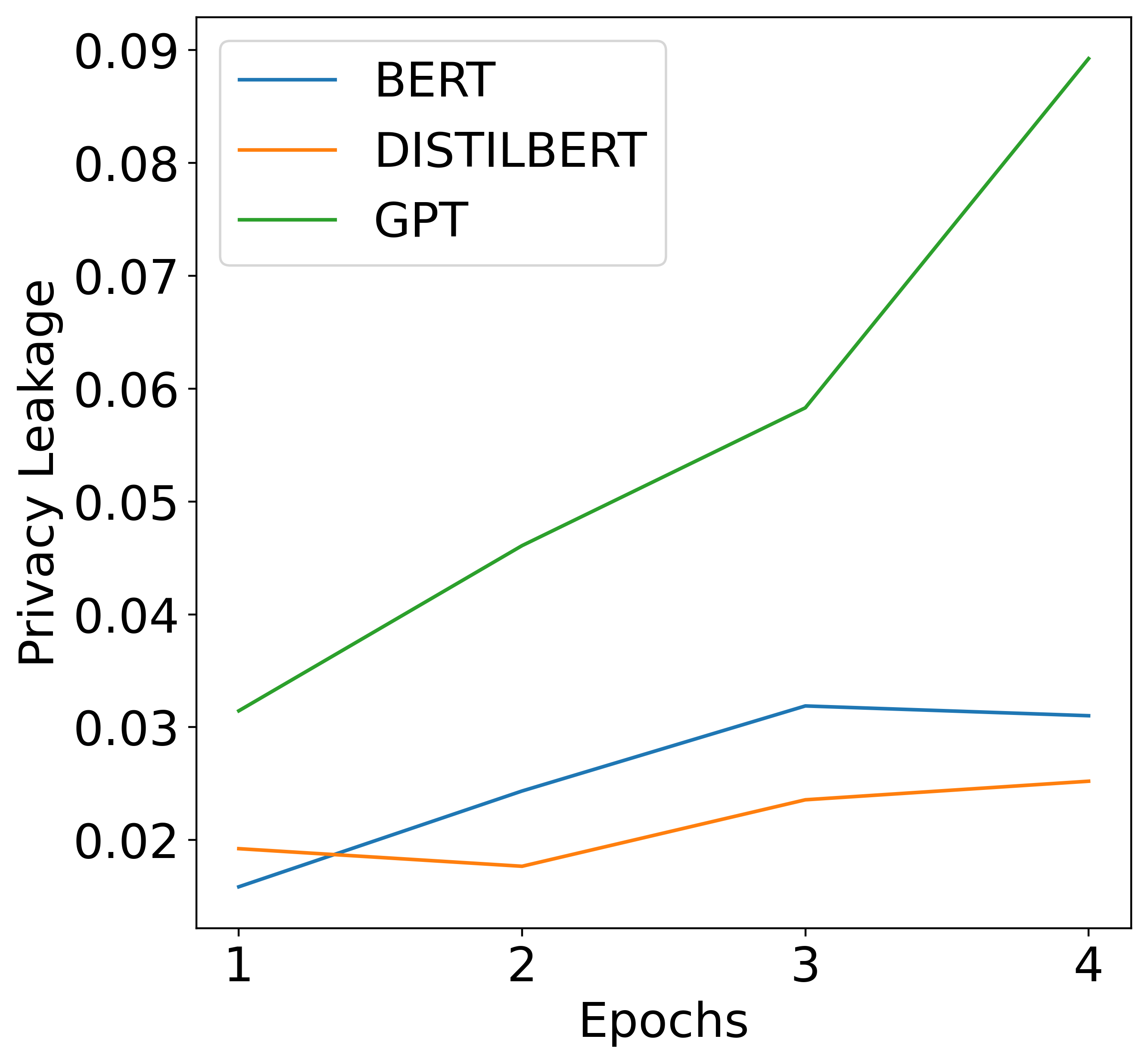}
    \caption{Black box attack privacy leakage for MIMIC-III data with patient level aggregate.}
    \label{fig:mimic_bb_patient_non_priv}
\end{figure}

%%%%%%%%% Graphs for Hospital 2 %%%%%%%%%%

\end{document}